\documentclass[pdflatex,sn-mathphys]{sn-jnl}

\usepackage{amsmath}
\usepackage{url}
\usepackage{lipsum}  
\usepackage{graphicx}
\usepackage{caption}
\usepackage{multirow}
\usepackage{multicol}
\usepackage{subcaption}
\usepackage{amssymb}
\usepackage{color}

\jyear{2022}%

\theoremstyle{thmstyleone}%
\theoremstyle{thmstyletwo}%

\theoremstyle{thmstylethree}%
\newtheorem{definition}{Definition}%

\begin{document}

\title[Driving Style Recognition]{Driving Style Recognition Using Interval Type-2 Fuzzy Inference System and Multiple Experts Decision-Making}


\author*[1]{\fnm{Iago} \sur{Pach\^eco Gomes}}\email{iagogomes@usp.br}

\author[1]{\fnm{Denis} \sur{Fernando Wolf}}\email{denis@icmc.usp.br}

\affil[1]{\orgdiv{University of S\~ao Paulo}, \orgdiv{Institute of Mathematics and Computer Science}, \orgname{Mobile Robotic Lab}, \city{S\~ao Carlos},  \state{S\~ao Paulo}, \country{Brazil} }


\abstract{
Driving styles summarize different driving behaviors that reflect in the movements of the vehicles. These behaviors may indicate a tendency to perform riskier maneuvers, consume more fuel or energy, break traffic rules, or drive carefully. Therefore, this paper presents a driving style recognition using Interval Type-2 Fuzzy Inference System with Multiple Experts Decision-Making for classifying drivers into calm, moderate and aggressive. This system receives as input features longitudinal and lateral kinematic parameters of the vehicle motion. The type-2 fuzzy sets are more robust than type-1 fuzzy sets when handling noisy data, because their membership function are also fuzzy sets. In addition, a multiple experts approach can reduce the bias and imprecision while building the fuzzy rulebase, which stores the knowledge of the fuzzy system. The proposed approach was evaluated using descriptive statistics analysis, and compared with clustering algorithms and a type-1 fuzzy inference system. The results show the tendency to associate lower and consistent kinematic profiles for the driving styles classified with the type-2 fuzzy inference system when compared to other algorithms, which is also in line with the more conservative approach adopted in the aggregation of the experts' opinions.
}

\keywords{Driving Style Recognition,  Type-2 Fuzzy Inference System, Interval Type-2 Fuzzy Sets, Fuzzy Multiple-Experts Decision-Making}

\maketitle

\section{Introduction}\label{sec1}
\label{sec:introduction}


When driving a vehicle on the road, a driver exhibits several specific behaviors that characterize a more general driving style. These behaviors are related to a variety of factors, including traffic conditions, the driver's mood, fatigue, driving experience, demographic characteristics such as gender and age, and many others \cite{ishibashi2007indices, lin2014overview, sagberg2015review, martinez2017driving}. Therefore, driving style defines the way drivers make decisions while driving in terms of maneuver choices and the way they execute them \cite{ishibashi2007indices,sagberg2015review,qi2015leveraging,huang2019safe}. According to Martinez et al. \cite{martinez2017driving} and Lv et al. \cite{lv2018driving}, they influence the control of the steering wheel, throttle pedal, brake pedal and other peripherals of the vehicle.


As Sagberg et al. \cite{sagberg2015review} notes, driving style recognition is an important task for many fields. In Advanced Driver-Assistance System (ADAS), knowledge of driving style can be used to either adapt vehicle control to driver preferences or to predict behaviors and take safety measures \cite{martinez2017driving,suzdaleva2018online,han2018statistical,dorr2014online}. In autonomous systems, it can be used to predict the driver's intentions and trajectory \cite{xing2019personalized, huang2019safe}. A well-known application is fuel economy, where the classifier divides driving styles into classes that save more fuel and classes that consume more fuel. Similarly, for electric vehicles, it is possible to classify driving styles according to energy consumption \cite{dorr2014online}. Another application is adaptive car insurance, where insurance companies adjust the cost of car insurance according to the risk drivers take based on their driving style \cite{sagberg2015review,zylius2017investigation}.


Due to the wide range of applications, there are also many techniques that can be used for classification, and also different driving style classes that are defined according to the needs of each application. For example, in adaptive vehicle insurance, driving style can be classified as cautious, average, expert, and reckless. The first class (cautious) describes drivers who put safety first when driving, anticipate the behavior of other vehicles, and always perform low-risk maneuvers. The expert driver, on the other hand, can control the vehicle with high precision in many traffic situations, even at high speeds. The reckless driver behaves carelessly and performs unpredictable or risky maneuvers \cite{lin2014overview}.


In an ADAS application, it is possible to classify driving styles into comfortable, normal, and sporty, which perform maneuvers and other actions with different kinematic profiles, such as velocity, acceleration, and jerk \cite{wang2011review,dorr2014online}. For safety applications, prediction of driver intentions and trajectories, and energy or fuel savings, it is also possible to classify driving styles as calm, moderate, and aggressive. A calm driver typically drives the vehicle to conserve fuel and performs actions or maneuvers in a conservative and safe manner with low speed, acceleration, or jerk profiles. The aggressive driver typically has higher fuel consumption and performs sharp and risky maneuvers and also drives with higher speed, acceleration, and jerk profiles. A normal or moderate driver, on the other hand, behaves between the two extremes (i.e., calm and aggressive), with average kinematic profiles \cite{wang2011review,huang2019safe,martinez2017driving,aljaafreh2012driving,lv2018driving,meiring2015review}.


According to Han et al. \cite{han2018statistical}, Liu et al. \cite{liu2019research}, and Martinez et al. \cite{martinez2017driving}, there are a number of challenges in driving style recognition due to the different applications, driving patterns, feature engineering, data noise, uncertainties, subjectivity in defining driving style classes, and validation of the model. In general, driving style recognition techniques are divided into model-based, learning-based, and rule-based. In model-based detection, a model represents each driving style, using techniques such as Hidden Markov Models and other stochastic models. In the learning-based method, machine learning techniques learns the driving style pattern from the data. Finally,  the rule-based methods represent the knowledge of experts.


The size of the observation time window for feature selection is another important concern in driving style recognition because this task is time dependent. A large time window, e.g., days or weeks, could therefore detect more consistent driving styles over time and is best suited for applications such as adaptive vehicle insurance. However, for real-time applications, such as intent or trajectory prediction, the size of the time window is limited to a few seconds. In such cases, the system is able to detect transient driving styles, i.e., a driver can adopt different driving styles on a route, and each detection is valid for a few more seconds \cite{murphey2009driver,meiring2015review}.


In this paper, we present a framework for driving style recognition that uses the Mamdani Interval Type-2 Fuzzy Inference System and Multiple Experts Decision-Making.  A Fuzzy Inference System (FIS) is a well-known rule-based technique that can deal with subjectivity and uncertainty. However, a standard type-1 fuzzy system has flaws in dealing with uncertainty and subjectivity because type-1 fuzzy sets have membership functions whose outputs are crisp values. Using type-2 fuzzy sets overcomes these weaknesses because the membership function is also fuzzy \cite{castillo2007type,chen2013fuzzy,melin2013review}. In addition, a Multiple Expert Decision-Making, or Group Decision-Making, is more robust to expert bias in defining system rules and also more robust to uncertainty and subjectivity  \cite{vanivcek2009fuzzy,chen2010fuzzy}.


The remainder of this paper is organized as follows: Section \ref{sec:related_works} presents some related work on driving style recognition; Section \ref{sec:fuzzy_type_2} discusses Interval Type-2 Fuzzy Sets, Type-2 Fuzzy Inference System, and Fuzzy Multiple Experts Decision-Making; Section \ref{sec:driving_style_recognition} describes the proposed driving style recognition; Section \ref{sec:results} presents and discusses the experimental results; and finally, Section \ref{sec:conclusion} presents remarks and future work.

\section{Related Works}
\label{sec:related_works}

As mentioned in the previous section, there are many applications for driving style, so there is no standard for defining classes (e.g., calm and aggressive), and different techniques can be used as well. In the absence of a large and labeled database of driving styles, most of the literature uses either unsupervised machine learning techniques, statistical analysis, or rule-based methods. Table \ref{tab:related_works} shows some related works with the techniques and driving style classes used.

\begin{table}[!bt]
    \caption{Primary Studies on Driving Style Recognition}
    \label{tab:related_works}
    \resizebox{\textwidth}{!}{
\scriptsize
\centering
      \begin{tabular}{ccc}
\hline
\textbf{Primary Studies}  & \textbf{Driving Style Classes}& \textbf{Techniques}\\

\hline
Johnson and Trivedi \cite{johnson2011driving}& \begin{tabular}[c]{@{}c@{}}Non-Aggressive\\ Aggressive\\\end{tabular} & Dynamic Time Warping      \\ 

\hline
\multirow{4}{*}{Han et al. \cite{han2018statistical}}  & \multirow{4}{*}{\begin{tabular}[c]{@{}c@{}}Normal\\ Aggressive\end{tabular}}& \multirow{4}{*}{\begin{tabular}[c]{@{}c@{}}Kernel Density Estimation,\\ Bayesian Decision,\\ Euclidean Decision,\\Fuzzy Inference System\end{tabular}}      \\
&     &  \\
&     &  \\ 
&     &  \\ 
\hline
\multirow{3}{*}{Xing, Lv, and Cao \cite{xing2019personalized}} & \multirow{3}{*}{\begin{tabular}[c]{@{}c@{}}Conservative\\ Moderate\\ Aggressive\end{tabular}}      & \multirow{3}{*}{\begin{tabular}[c]{@{}c@{}}Gaussian Mixture Model\\ Clustering\end{tabular}}      \\
&     &  \\
&     &  \\ 
\hline
\multirow{3}{*}{Qi et al. \cite{qi2015leveraging}}     & \multirow{3}{*}{\begin{tabular}[c]{@{}c@{}}Cautious\\ Moderate\\ Aggressive\end{tabular}}  & \multirow{3}{*}{\begin{tabular}[c]{@{}c@{}}Ensemble Clustering Method\\ based on Kernel-Fuzzy C-means,\\ Latent Dirichlet Allocation\end{tabular}}\\
&     &  \\
&     &  \\ 
\hline
\multirow{4}{*}{Suzdaleva and Nagy \cite{suzdaleva2018online}} & \multirow{4}{*}{\begin{tabular}[c]{@{}c@{}}Below Normal\\ Normal\\ Aggressive\\ Very Aggressive\end{tabular}}& \multirow{4}{*}{Fuzzy Inference System}     \\
&     &  \\
&     &  \\
&     &  \\ 
\hline
\multirow{3}{*}{\begin{tabular}[c]{@{}c@{}}Dörr, Grabengiesser\\ and Gauterin \cite{dorr2014online}\end{tabular}}& \multirow{3}{*}{\begin{tabular}[c]{@{}c@{}}Comfortable\\ Normal\\ Sporty\end{tabular}}     & \multirow{3}{*}{Fuzzy Inference System}     \\
&     &  \\
&     &  \\ 
\hline
\multirow{3}{*}{\begin{tabular}[c]{@{}c@{}}Murphey, Milton\\ and Kiliaris \cite{murphey2009driver}\end{tabular}} & \multirow{3}{*}{\begin{tabular}[c]{@{}c@{}}Calm\\ Normal\\ Aggressive\end{tabular}} & \multirow{3}{*}{Threshold-based Analysis}   \\
&     &  \\
&     &  \\ 
\hline
\multirow{5}{*}{\begin{tabular}[c]{@{}c@{}}Constantinescu, Marinoiu\\ and Vladoiu \cite{constantinescu2010driving}\end{tabular}} & \multirow{5}{*}{\begin{tabular}[c]{@{}c@{}}Non-Aggressive\\ Somewhat Non-Aggressive\\ Neutral\\ Moderately Aggressive\\ Very Aggressive\end{tabular}} & \multirow{5}{*}{\begin{tabular}[c]{@{}c@{}}Hierarchical Cluster Analysis,\\ Principal Component Analysis\end{tabular}}   \\
&     &  \\
&     &  \\
&     &  \\
&     &  \\
\hline
\multirow{3}{*}{Liu et al. \cite{liu2019research}}     & \multirow{3}{*}{\begin{tabular}[c]{@{}c@{}}Conservative\\ General\\ Aggressive\end{tabular}}       & \multirow{3}{*}{\begin{tabular}[c]{@{}c@{}}Principal Component Analysis,\\ Fuzzy C-means,\\ Support Vector Machine\end{tabular}}   \\
&     &  \\
&     &  \\ 
\hline
\multirow{3}{*}{Ma et al. \cite{ma2021driving}}& \multirow{3}{*}{\begin{tabular}[c]{@{}c@{}}Cautious\\ Normal\\ Aggressive\end{tabular}}    & \multirow{3}{*}{\begin{tabular}[c]{@{}c@{}}Threshold-based Analysis,\\ Principal Component Analysis,\\ K-Means Clustering\end{tabular}}    \\
&     &  \\
&     &  \\
\hline
\multirow{2}{*}{Tian et al. \cite{tian2019adaptive}}   & \multirow{2}{*}{\begin{tabular}[c]{@{}c@{}}Conservative\\ Aggressive\end{tabular}}  & \multirow{2}{*}{\begin{tabular}[c]{@{}c@{}}K-Nearest Neighbors,\\ Expectation-Maximization\end{tabular}}  \\
&     &  \\ 
\hline
\multirow{4}{*}{Guo et al. \cite{guo2019adaptive}}     & \multirow{4}{*}{\begin{tabular}[c]{@{}c@{}}Economical\\ Soft\\ Normal\\ Aggressive\end{tabular}}   & \multirow{4}{*}{Fuzzy Inference System}     \\
&     &  \\
&     &  \\
&     &  \\ 
\hline
\multirow{5}{*}{Bejani and Ghatee \cite{bejani2018context}}    & \multirow{5}{*}{\begin{tabular}[c]{@{}c@{}}Low Risk\\ Normal Risk\\ High Risk\\ Very High Risk\end{tabular}} & \multirow{5}{*}{\begin{tabular}[c]{@{}c@{}}Decision-Tree,\\ Support Vector Machine,\\ Multi-Layer Perceptron,\\ K-Nearest Neighbors,\\ Fuzzy Inference System\end{tabular}} \\
&     &  \\
&     &  \\
&     &  \\
&     &  \\ 
\hline
\multirow{3}{*}{\begin{tabular}[c]{@{}c@{}}Mohammadnazar, Arvin\\ and Khattak \cite{mohammadnazar2021classifying}\end{tabular}}  & \multirow{3}{*}{\begin{tabular}[c]{@{}c@{}}Calm\\ Normal\\ Aggressive\end{tabular}} & \multirow{3}{*}{\begin{tabular}[c]{@{}c@{}}K-Means,\\ K-medoid\end{tabular}}    \\
&     &  \\
&     &  \\ 
\hline
\multirow{3}{*}{De Rango et al. \cite{de2022fuzzy}}    & \multirow{3}{*}{\begin{tabular}[c]{@{}c@{}}Normal\\ Aggressive\\ Very Aggressive\end{tabular}}     & \multirow{3}{*}{Fuzzy Inference System}     \\
&     &  \\
&     &  \\ 
\hline
\multirow{5}{*}{Brombacher et al. \cite{brombacher2017driving}}        & \multirow{5}{*}{\begin{tabular}[c]{@{}c@{}}Very Defensive\\ Defensive\\ Normal\\ Sporty\\ Very Sporty\end{tabular}}    & \multirow{5}{*}{\begin{tabular}[c]{@{}c@{}}Multi-Layer Perceptron,\\ Fuzzy Sets\end{tabular}}   \\
 &      & \\
 &      & \\
 &      & \\
 &      & \\ 
 \hline
 
 \multirow{2}{*}{Wang et al. \cite{wang2017driving}}  & \multirow{2}{*}{\begin{tabular}[c]{@{}c@{}}Normal\\ Aggressive\end{tabular}}     & \multirow{2}{*}{\begin{tabular}[c]{@{}c@{}}Semisupervised Support Vector Machine,\\ K-Means\end{tabular}}        \\
 &      & \\ 
 \hline
 \multirow{3}{*}{Cordero et al. \cite{cordero2020recognition}}& \multirow{3}{*}{\begin{tabular}[c]{@{}c@{}}Ecological\\ Normal\\ Aggressive\end{tabular}}     & \multirow{3}{*}{Multi-Layer Fuzzy Inference System}         \\
&     &     \\
&     &     \\ 
\hline
\multirow{3}{*}{Deng et al. \cite{deng2017driving}}           & \multirow{3}{*}{\begin{tabular}[c]{@{}c@{}}Mild\\ Moderate\\ Aggressive\end{tabular}}          & \multirow{3}{*}{Hidden Markov Model}  \\
&     &        \\
&     &        \\ 
\hline
\end{tabular}
    }
    
     {\vspace{1em} \scriptsize \begin{tabular}{@{}l@{}}* each cluster measures the level of fuel consumption from the lower to the higher \\level.\end{tabular}  }
     
\end{table}

\subsection{Model-based and Supervised Driving Style Recognition}

One approach to detecting driving styles is to design an analytical model for each driving style, using mathematical and/or statistical tools. For example, threshold analysis can distinguish different styles based on a metric (e.g., distance function) and known bounds for each class. In addition, probabilistic graph models can also learn to represent individual patterns of styles, such as Hidden Markov Models (HMM) and Gaussian Processes (GP).

In Johnson and Trivedi \cite{johnson2011driving}, a detection of driving maneuvers with a distinction of driving style between `non-aggressive' and `aggressive' for each maneuver was developed using Dynamic Time Warping (DTW) and a smartphone-based sensor fusion (accelerometer, gyroscope, magnetometer, GPS, and video) that records potentially aggressive events. The drawbacks of the proposed method are the calibration of the smartphone sensors, including the positioning of the smartphone in the vehicle, and the sensitivity of the approach to the noise of the sensors, which is minimized only by a low-pass filter. Similarly, Deng et al. \cite{deng2017driving} proposed a braking analysis method using HMM that classifies drivers into `mild', `moderate', and `aggressive'. They evaluated their system using data extracted from the Controller Area Network (CAN) bus of vehicles that drove on an expressway. There were a total of 30 human drivers with different genders, ages, and driving experience. Moreover, each driver self-reported their style after the experiment.

Murphey, Milton, and Kiliaris \cite{murphey2009driver} presented a driving style classification based on the analysis of the vehicle's jerk profile, classifying segments of the track into `calm', `normal', and `aggressive' driving styles. The algorithm calculates the ratio between the standard deviation of the jerk profile within a time window, and the mean value of the jerk on a given road type (e.g., highway and ramps). A threshold-based analysis performs the classification. The authors compared the performance of the system with time windows of sizes $15$, $9$, $6$, and $3$ seconds, with $9$ and $6$ yielding the best results. The disadvantages of the method are the calibration of the thresholds, which must also be robust to noise, the use of only one feature to characterize a driving style, the use of only longitudinal features, and the difficulty of characterizing an average driving style for a given road type by using only the mean of the jerk profile. 

There are also primary studies in the literature that apply supervised machine learning techniques to detect driving styles, in cases where the available data have annotations of each target driving style. Multi-Layer Perceptron (MLP) and Support Vector Machines (SVM) are the most commonly used techniques. Bejani and Ghatee \cite{bejani2018context} presented an ensemble learning algorithm that combines MLP, SVM with a radial kernel, and K-Nearest Neighbors (KNN) to classify the risk level of maneuvers (i.e., U-Turn, Turn, and Lane-Change) from low to very risky. They evaluated their system using data from the magnetometer of smartphones inside vehicles of 27 drivers,  where two experts annotated the data when traveling as passengers.  Moreover, they applied two different Fuzzy Inference Systems (FIS) to aggregate the results and knowledge of the driver profile into the final driving style.

Altenatively, Brombacher et al. \cite{brombacher2017driving} developed a detection framework using MLP with fuzzy outputs. They used data from an embedded system inside each vehicle (i.e., Raspberry Pi) that extracted longitudinal and lateral acceleration, yaw rate, and velocity. Moreover, they applied a Savitzky-Golay smoothing filter (polynomial degree 2 and a window size of 9 samples) to reduce the noise in the data. The output of the model is a membership vector that indicates the probability for each driving style. 
Liu et al. \cite{liu2019research} applied PCA, Fuzzy C-Means (FCM) clustering, and SVM to data from $51$ drivers, and classified them as `conservative,' `general,' and `aggressive.' Each driver drove a $25$ km route with an average time of $40$ min. The proposed approach analyzed the longitudinal features of the routes (i.e., velocity, longitudinal acceleration, and throttle) after discretizing them using information entropy. The PCA reduced the dimension of the feature vector created with statistical features (i.e., mean and standard deviation) of the data.

\subsection{Clustering-based Driving Style Recognition}

In the absence of labeled data, primary studies on driving style recognition focus on the use of unsupervised learning algorithms, such as clustering, and expert systems. The purpose of clustering algorithms is to partition the feature space by separating it into groups that share similarities  \cite{xu2015comprehensive}. Thus, assuming that each driving style produces a distinct kinematic and dynamic profile of motion, it is possible to partition the feature space into clusters and classify each style based on a similarity metric (e.g., distance function). 

Xing, Lv, and Cao \cite{xing2019personalized} presented an intention-aware trajectory prediction based on driving style detection, where a Gaussian Mixture Model (GMM) clustering classifies driving styles into `conservative', `moderate' and `aggressive'. For each class, a dedicated  Multi-layer Perceptron (MLP) performs the regression of the predicted trajectory, also using the features extracted from a Long Short-Term Memory (LSTM). This approach was validated using the NGSIM dataset, which is known to contain significant noise data with impossible kinematic parameters. To exploit this data, the authors used an interpolated curve method and a low-pass filter to reduce the impact of noisy data. In Qi et al. \cite{qi2015leveraging}, an Ensemble Clustering Method (ECM) based on Kernel Fuzzy C-means (K-FCM) and modified Latent Dirichlet Allocation (LDA) was applied to detect driving style by analyzing longitudinal features of vehicle kinematics (i.e., velocity and acceleration) and the time gap between the leading vehicle and the trailing vehicle (target) in a car-following scenario.

Suzdaleva and Nagy \cite{suzdaleva2018online} developed a recursive Bayesian mixture-based cluster analysis used in a fuel economy application. This method identifies the level of fuel consumption taking into account velocity, throttle pedal position, and gear. The recursiveness of the application occurs when the classification at the current time frame depends on the driving style detected at an earlier time frame for the same vehicle. Constantinescu, Marinoiu and Vladoiu \cite{constantinescu2010driving} performed a statistical analysis using Principal Component Analysis (PCA) and Hierarchical Cluster Analysis (HCA) to classify drivers into driving styles that are more likely to commit traffic violations or be involved in accidents. The method analyzed data from $23$ different drivers collected within $2$ to $5$ working days using statistical features of kinematic parameters such as mean and standard deviation of velocity, acceleration, positive acceleration, braking (negative acceleration), and mechanical work. The time window of days allowed the authors to identify time-consistent driving styles rather than transient driving styles that are the target of real-time applications.

Ma et al. \cite{ma2021driving} proposed a framework that used K-Means clustering to classify driving styles for online car-hailing platforms. They gathered data from 10 drivers (9 males and 1 female) of different ages on a single working day on routes with similar weather and pavement conditions to reduce disturbance factors. To further reduce the noise in the data, they smoothed the data using a time window filter and removed outliers. Alternatively, Tian et al. \cite{tian2019adaptive} combined KNN with Expectation-Maximization (EM) to find the closest samples of a given feature vector based on their similarity, and to estimate the probabilities of the driving styles using the Bayes theorem. Mohammadnazar, Arvin and Khattak \cite{mohammadnazar2021classifying} compared K-Means and K-medoid clustering in the context of connected vehicles for driving styles classification taking into account the driving environment (i.e., highway, commercial, and residential streets) and the volatile of their kinematic parameters. Wang et al. \cite{wang2017driving}, in turn, proposed a semisupervised SVM to classify the longitudinal behavior of drivers when driving on curvy roads. This method uses KMeans to pre-label part of the data, and then combine the labeled and unlabeled data in training an SVM. Moreover, they evaluated their system using data collected from 20 drivers in a stationary driving simulator.

\subsection{Knowledge-based Driving Style Recognition}

In addition to the aforementioned learning-based, model-based, and statistical methods for detecting driving styles, there are also rule-based techniques that are able to represent expert knowledge using rules, as shown in Eq. \ref{eq:rule}. In this sense, the primary studies often use Fuzzy Inference System (FIS).  For instance, Aljaafreh, Alshabatat and Al-Din \cite{aljaafreh2012driving} proposed a FIS for classifying driving style based on longitudinal features detected by a 2-axis accelerometer into `below normal', `normal', `aggressive', and `very aggressive' categories. The input to the system is the Euclidean norm of lateral and longitudinal acceleration, filtered by a moving average filter, and velocity. 

\begin{eqnarray}
    \label{eq:rule}
    \mathbf{if}\;\mathit{logical\;preposition}\;\mathbf{then}\;\mathit{logical\;consequence}
\end{eqnarray}

Similarly,  Dörr, Grabengiesser, and Gauterin \cite{dorr2014online} also used an FIS to detect driving style. Their system extracted input features from the vehicle's CAN bus and a road class (dirt track, urban street, rural roads, and freeway) from the navigation system. Each road class has a specific FIS, which also used different input parameters (e.g., acceleration, velocity, deceleration, time gap). The authors also used a Kalman Filter (KF) to reduce noise in the data.  Han et al. \cite{han2018statistical}, in turn, proposed a statistic-based detection of driving style between `normal' and `aggressive' by modeling the uncertainty of driver behavior with Kernel Density Estimation (KDE), and classifying driving styles with a modified Bayesian decision using Euclidean distance. They also developed a Fuzzy Inference System (FIS) to compare the approaches. The disadvantages of this system are the scalability of the number of driving style classes, and the use of only longitudinal features (i.e., vehicle speed, throttle opening, and longitudinal acceleration) of vehicle motion.

According to Guo et al. \cite{guo2019adaptive}, fuzzy logic is very suitable for driving style recognition because there are no clear boundaries between the classes. Therefore, they proposed a fuzzy system to evaluate the accelerator pedal opening and its change rate in order to differentiate a driver into `economical', `soft', `normal', and `aggressive' regarding energy consumption. De Rango et al. \cite{de2022fuzzy} developed a hierarchical FIS, which took the speed, acceleration, environment context, and jerk profile as input.  This approach stacks different FIS in order to estimate the final output. In this case, the first system is responsible for identifying the enviroment context (urban, sub-urban, and extra-urban), and the final system for classifying the driving style (normal, aggressive, and very aggressive). Cordero et al. \cite{cordero2020recognition} also employed an hierarchical FIS. However, they stacked three models responsible for, respectively, recognizing the driver emotion (anger, happy, sad, fear, surprise, and neutral), state (relaxed, wakefulness, stressed, pleasant, sleepy, and fatigue), and driving style (ecological, normal, and aggressive). To evaluate their approach, they created an artificial dataset using real data from different projects to satisfy all required input descriptors in each FIS.

This paper also proposes a Fuzzy Inference System for driving style detection. However, instead of standard type-1 fuzzy sets, we use Interval type-2 fuzzy sets \cite{chen2013fuzzy}, which are more robust to uncertainty and subjectivity inherent to the rule-based reasoning systems. Moreover, the rules of the system were created using Fuzzy Multiple Experts Decision-Making, with a type-1 fuzzy aggregation operator, known as Ordered Weighted Averaging (OWA), which also reduces the bias of the experts in creating the rules of the system \cite{vanivcek2009fuzzy,chen2010fuzzy,zhou2008type}. The proposed system was validated with the publicly available large dataset Argoverse $v.1$ \cite{chang2019argoverse}, and compared with clustering algorithms (K-Means, GMM clustering, Fuzzy C-means, and Agglomerative Hierarchical Clustering) using descriptive statistical analysis on the clusters. The input combines statistical features (mean and standard deviation) of the longitudinal (velocity, acceleration, and deceleration) and lateral (lateral jerk) kinematic parameters of the vehicle motion over a time window. All input features are derivative of the  space (position of the vehicle). To estimate their values and reduce noise, the proposed approach evaluates the use of Extended Kalman Filter (EKF) and Savitzky-Golay filter.


\section{Interval Type-2 Fuzzy Logic System}
\label{sec:fuzzy_type_2}

Fuzzy logic is a mathematical tool capable of approximating human reasoning in a decision-making process characterized by imprecision, uncertainty, subjectivity, conflicting information, and partially true or possible information \cite{zadeh2008there,valavskova2014role}. The uncertainties and inaccuracies in the decision process have various sources \cite{castillo2007type,melin2013review}, which are measurement or observation; process; model; estimation; or, implementation. Some of these sources may be related to computational approximation, sensor noise, subjectivity in defining variables, incorrect or incomplete specification of model behavior, stochastic variables, and randomness of system dynamics. Therefore, a fuzzy logic system handles most of these uncertainties by using fuzzy sets, which can be represented by linguistic terms, with a degree of uncertainty within the range $[0,1]$, defined by a membership function for each fuzzy set. 

There are two types of fuzzy sets, type-1 fuzzy sets whose membership functions have crisp outputs, and, type-2 fuzzy sets whose membership functions are outputs are type-1 fuzzy sets \cite{castillo2007type}. According to Melin and Castillo \cite{melin2013review}, when using a type-1 fuzzy set, we assign a certainty to a fuzzy set because the output of its membership function is crisp. However, in systems dealing with uncertainty, it is not recommended to use this approach. In this sense, type-2 fuzzy sets are advisable because their membership functions also represent uncertainties \cite{karnik1999type,castillo2007type}.

This section describes the definition of a type-2 fuzzy set,  interval type-2 fuzzy set, and the parameters of a trapezoidal membership function (see subsection \ref{subsec:fuzzy2:fuzzy_sets})  \cite{chen2013fuzzy, castillo2007type,melin2013review,wu2019recommendations};  the components of a type-2 Fuzzy Inference System (see subsection \ref{subsec:fuzzy2:fls}) \cite{wu2019recommendations, karnik1999type, chen2013fuzzy}; and, the fuzzy aggregate operation, called  Ordered Weighted Averaging (OWA), applied in the context of Fuzzy Multiple Experts Decision-Making (see subsection \ref{subsec:fuzzy2:multiple_experts_dm}) \cite{vanivcek2009fuzzy,zhou2008type,chen2010fuzzy}.

\subsection{Type-2 Fuzzy Sets}
\label{subsec:fuzzy2:fuzzy_sets}

\begin{definition}[Type-2 Fuzzy Set]
    A type-2 fuzzy set, denoted by $\tilde{A} \in \mathit{U}$, in which $U$ is the universe of discourse, is a fuzzy set whose membership function is a type-2 membership function, $\mu_{\tilde{A}}$ , shown as follows: 
    
    \begin{eqnarray}
        \tilde{A} = \left\{\left(\left(x, u\right), \mu_{\tilde{A}}\left(x,u\right)\right) \mid \forall_{x} \in \mathit{U}, \forall_{u} \in \mathit{J}_{x} \subseteq \left[0,1\right]\right\}
    \end{eqnarray}
    in which, $0 \leq \mu_{\tilde{A}}\left(x,u\right) \leq 1$. $J_x$ is known as the primary membership of $x$, and $\mu_{\tilde{A}}\left(x,u\right)$ is the secondary set, and a type-1 fuzzy set.
\end{definition}
\vspace{1em}

\begin{definition}[Interval Type-2 Fuzzy Set]
    Let $\tilde{A}$ be a type-2 fuzzy set in the universe of discourse $\mathit{U}$, and $\mu_{\tilde{A}}$ its type-2 membership function. Then,  $\tilde{A}$ is a interval type-2 fuzzy set if and only if:
    \begin{eqnarray}
    \mu_{\tilde{A}}\left(x,u\right) = 1, \quad \forall_{u} \in \mathit{J}_{x} \subseteq \left[0,1\right]
    \end{eqnarray}
    
    Then, the type-2 membership function, $\mu_{\tilde{A}}\left(x,u\right)$, can be expressed by a type-1 inferior membership function and a type-1 superior membership function, as follows:
    \begin{eqnarray}
       \tilde{A}=\left\{((x, u), 1) \mid \forall_{x} \in \mathit{U}, \forall_{u} \in \left[\underline{\mu}_{A}(x),\bar{\mu}_{A}(x)\right] \subseteq [0,1]\right\}
    \end{eqnarray}
    
\end{definition}

\begin{figure}[!tb]
      \centering
      \includegraphics[width=0.45\textwidth]{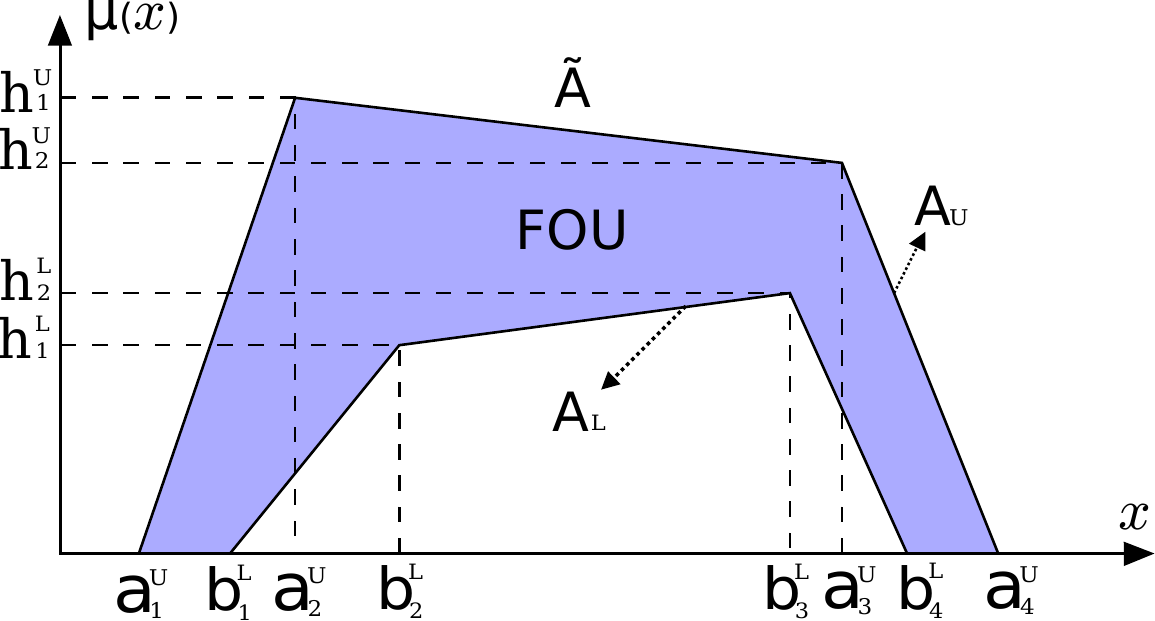}
      \caption{Type-2 Trapezoidal Membership Function}
      \caption*{\scriptsize Source: based on Chen and Wang \cite{chen2013fuzzy}.}
      \label{fig:trapmf}
  \end{figure}

\vspace{1em}
\begin{definition}[Trapezoidal Membership Function]
\label{def:fuzzy:trapmf}
An interval type-2 trapezoidal set, denoted by $\tilde{A}$, can be represented by an upper and lower type-1 membership function, as $\tilde{A}=\left({A}_{U}, A_{L}\right)$, in which $A_{U}$ and $A_{L}$ are, respectively, the upper and lower membership functions, and can be parameterized as follows: 
  \begin{eqnarray}
      A_{U} &=& \left(a_{1}^{U},a_{2}^{U},a_{3}^{U},a_{4}^{U}, h_{1}^{U},h_{2}^{U}\right)\\
      A_{L} &=& \left(b_{1}^{L},b_{2}^{L},b_{3}^{L},b_{4}^{L}, h_{1}^{L}, h_{2}^{L}\right)
  \end{eqnarray}
  in which $A_{L} \subseteq A_{U}$,  $h_{1}^{U} \in \left[0,1\right]$, $h_{2}^{U} \in \left[0,1\right]$, $h_{1}^{L} \in \left[0,1\right]$, and $h_{2}^{L} \in \left[0,1\right]$.
  
  When $h_{1}^{U}=h_{2}^{U}$ and $h_{1}^{L}=h_{2}^{L}$, then $A_{U} = \left(a_{1}^{U},a_{2}^{U},a_{3}^{U},a_{4}^{U}, h^{U}\right)$ and $A_{L} = \left(a_{1}^{L},a_{2}^{L},a_{3}^{L},a_{4}^{L}, h^{L}\right)$. Figure \ref{fig:trapmf} shows the graphical representation of the type-2 generalized trapezoidal membership function.  The area between the upper and lower membership functions is called the Footprint Of Uncertainty (FOU), and represents the uncertainty of the type-2 fuzzy set, that is, the smaller the area, the less uncertainty \cite{mendel2002type}.


\end{definition}

\subsection{Type-2 Fuzzy Inference System}
\label{subsec:fuzzy2:fls}

\begin{figure}[!tb]
      \centering
      \includegraphics[width=0.6\textwidth]{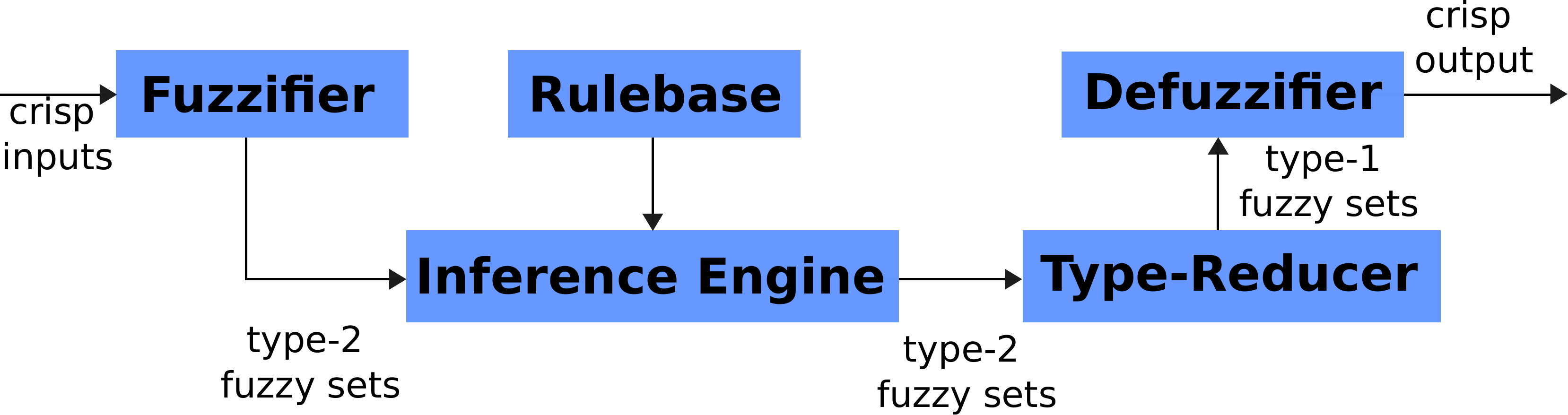}
      \caption{Type-2 Fuzzy Inference System}
      \caption*{\scriptsize Source: based on Wu and Mendel \cite{wu2019recommendations}, and  Karnik, Mendel and Liang \cite{karnik1999type}.}
      \label{fig:fis_t2}
  \end{figure}

A Fuzzy Inference System (FIS) is a rule-based technique that converts input data into fuzzy sets and uses experts knowledge, expressed in logical prepositions, to make decisions \cite{castillo2007type}. The decision process is performed by a fuzzy inference engine built on the basis of fuzzy logic \cite{wu2019recommendations,karnik1999type}. According to Melin and Castillo \cite{melin2013review}, the structure of a type-2 FIS (T2-FIS) is similar to that of a type-1 FIS (T1-FIS), except for an additional component called the type-reducer, and the adaptation of components for processing type-2 fuzzy sets. Figure \ref{fig:fis_t2} shows a block diagram of a T2-FIS with the following components: \emph{fuzzifier}; \emph{rulebase}; \emph{inference engine}; \emph{type-reducer}; and, \emph{defuzzifier}.

\begin{itemize}
    \item \textbf{Fuzzifier}: As stated in  Wu and Mendel \cite{wu2019recommendations}; and, Karnik, Mendel and Liang \cite{karnik1999type}, the first block of a T2-FIS maps an input vector of crisp values, $x=\left(x_{1}, \dots, x_{n}\right),  x \in \mathbb{R}^{n}$, to type-2 fuzzy sets, called antecessors,  representing the input of the \emph{fuzzy inference engine}, as follows: 
    \begin{eqnarray}
        \tilde{a} = \left\{\tilde{a}_1, \dots, \tilde{a}_n\right\}, \quad \tilde{a}_1 \in \tilde{A}_1, \dots, \tilde{a}_n \in \tilde{A}_n 
    \end{eqnarray}
    \item \textbf{Rulebase}: This component stores the knowledge of the system,  which is usually defined by human experts.  The rules in a T2-FIS are created in the same way as in a T1-FIS, through logical proposition, such as: 
    \begin{eqnarray}
        \label{eq:fuzzy2:rules}
        &\mathbf{if}& (\tilde{a}_{1}~\mathbf{is}~\tilde{A}_{1}^{'})~\mathbf{and}~\dots~\mathbf{and}~(\tilde{a}_{n}~\mathbf{is}~\tilde{A}_{n}^{'}), \nonumber\\&\mathbf{then}&~\tilde{y}\left(\tilde{a}\right)~\mathbf{is}~\tilde{Y}^{'}
    \end{eqnarray}
    in which, $\tilde{y}\left(\tilde{a}\right)$ is the output of the T2-FIS, and belongs to the set of possible outputs $\tilde{Y}$, called consequent. 
    
    
    According to Melin and Castillo \cite{melin2013review}, one of the problems in building a fuzzy inference system is the number of membership functions of the input sets, since, as shown in Eq. \ref{eq:fuzzy2:rules}, the number of rules grows exponentially with the number of membership functions. The structure of rules given in Eq. \ref{eq:fuzzy2:rules}  is known as \emph{Zadeh}, whose outputs are type-2  fuzzy sets, and are found in fuzzy systems known as Mamdani Fuzzy Systems. There are also the rules of type \emph{ TSK } (Takagi-Sugeno-Kang) whose outputs are an interval with crisp bounds, estimated by linear combinations of the inputs.
    
    \item \textbf{Inference Engine}: As stated in  Castillo et al. \cite{castillo2007type}, the \emph{fuzzy inference engine} maps the input, $x$, into an output $\tilde{y}\left(x\right)$, through fuzzy logic reasoning, using the rules stored in the \emph{rulebase}. This mapping is considered a nonlinear partitioning of the input space $\tilde{A}_1 \times \dots \times \tilde{A}_n$ into an output $\tilde{y}\left(x\right)$ \cite{melin2013review}. According to  Wu and Mendel \cite{wu2019recommendations}, the most popular fuzzy operation for reasoning are the adaptation of minimum and product t-norms for type-2 fuzzy sets. 
    
    \item \textbf{Type-Reducer}: This component transforms the output of the \emph{fuzzy inference engine}, a type-2 fuzzy set, into a type-1 fuzzy set, using extended versions of type-1 defuzzification methods \cite{melin2013review}. According to  Wu and Mendel \cite{wu2019recommendations}, there are many methods to reduce fuzzy sets from type-2 to type-1, and the most commonly used is the \emph{center-of-sets}, which calculates centroids of type-2 sets (consequent) whose domain is discretized \cite{karnik1999type,wu2019recommendations}. 

    \item \textbf{Defuzzifier}: The output of the \emph{type-reducer} can be used as the final result of the T2-FIS, but it is also possible to obtain a crisp value as output through the defuzzification methods \cite{karnik1999type,wu2019recommendations}. As described in Karnik, Mendel, and Liang \cite{karnik1999type}, one of the techniques is to find the centroid of the type-reduced set. For interval sets of type 2, the centroid is the midpoint of the domain \cite{melin2013review}.
    
\end{itemize}

\subsection{Fuzzy Multiple Experts Decision-Making}
\label{subsec:fuzzy2:multiple_experts_dm}


The Fuzzy Inference System (FIS) is a rule-based technique capable of handling uncertainty and subjectivity by using fuzzy sets and fuzzy logic. One of the main components of FIS is the \emph{rulebase}, usually created by a human expert, which stores the knowledge of the system about the problem addressed. However, rules created by human experts can either carry the expert's bias, be hampered by any lack of knowledge and experience during the judgment, or be a result of a misjudgment. Therefore, according to Vaníček, Vrana and Aly \cite{vanivcek2009fuzzy}, to overcome these inaccuracies, it may be suitable to create the \emph{rulebase} using the opinion, or judgment, of several experts. This approach is known as Fuzzy Multiple Experts Decision-Making (FMEDM) or Fuzzy Group Decision-Making (FGDM).

An FMEDM system starts with getting experts' opinions, which are combined into either a single crisp value or a fuzzy set, using a fuzzy aggregation operator that collectively reflects all the individual preferences of the experts \cite{zhou2008type,capuano2017fuzzy}.
However, before the aggregation step, according to Vaníček, Vrana, and Aly \cite{vanivcek2009fuzzy}, the expert opinions in the same universe of discourse must be fuzzyified, which can be done by an increasing mapping function, with the domain $[0,1]$, of all possible fuzzy sets.


There are several aggregation operators for FMEDM using either type-1 or type-2 fuzzy sets. One well-known operator is Ordered Weighted Averaging (OWA) proposed by Yager \cite{yager1988ordered}, which combines preferences according to their content rather than their source, as in weighted averaging, that take into account the order of each expert's opinion in the input vector \cite{vanivcek2009fuzzy}. Moreover, the OWA method consists of three steps, starting with the descending order of expert opinions, then estimating the weights, and finally applying the aggregation function  \cite{zhou2008type}.


The OWA operator not only has a low computational cost, but is also versatile in that the choice of weights can change the behavior of the operator \cite{zhou2008type,capuano2017fuzzy}. Yager  \cite{yager1988ordered} also proposed a method for identifying the weight vector using linguistic quantifiers (e.g., 'most', 'almost all', 'at least 50 percent') by modeling the proportion that these linguistic quantifiers represent in the closed interval $\left[0,1\right]$.
Definition \ref{def:owa} summarizes the mathematical formulation of OWA, and Definition \ref{def:owa:q} shows how to estimate the weights.

\begin{definition}[Type-1 OWA Operator]
\label{def:owa}
The Ordered Weighted Averaging, denoted by $\phi$, is a mapping function, $\phi: [0,1]^{n} \to [0,1]$, which aggregates opinions of $n$ experts in the domain $[0,1]$, as follows:
\begin{eqnarray}
    \phi\left(a_{1}, \dots, a_{n}\right)&=&\sum_{i=1}^{n}w_{i}\cdot a_{\sigma\left(i\right)},\quad n\geq 2
\end{eqnarray}
in which, $a_{i}$ and $a_{\sigma\left(i\right)}$  are the experts' opinions, respectively, before and after descending ordering, and $i = \left\{1,2,\dots, n\right\}$.  In addition, $w_i$ is a component of the weight vector, $\vec{w}=\left(w_1, w_2, \dots, n\right) \in \left[0,1\right]^{n}$, where  $\sum^{m}_{i=1}w_{i}=1$.
\end{definition}

\begin{definition}[Type-1 OWA Linguistic Quantifier]
\label{def:owa:q}
The linguistic quantifier, denoted by $Q$, is a membership function, $Q: [0,1] \to [0,1]$, that indicates the degree to which a fuzzy set, $r$, satisfies the linguistic connotation desired for the aggregation function, such as 'most', 'almost all', and 'at least half'. This function fulfill the conditions, $Q\left(1\right)=1$, $Q\left(0\right)=0$, and it is monotonically non-decreasing, which means that if $x>y$ then $Q\left(x\right) \geq Q\left(y\right)$. 

Therefore, the definition of the  weights $w_i$ and the function $Q$ are as follows:
\begin{eqnarray}
    w_{i}&=&Q\left(\frac{i}{n}\right) - Q\left(\frac{i-1}{n}\right), \quad i=1,\dots,n\\\nonumber\\
     Q\left(r\right)&=&\begin{cases}0, & r < a\\\frac{r-a}{b-a}, & a \leq r \leq b\\1, & r > b\end{cases}
\end{eqnarray}
in which $n$ is the number of experts with $n\geq2$, $r \in [0,1]$ is an expert opinion, and $(a,b)$ are parameters of the function $Q$ that denotes the linguistic connotation, e.g., $\left(a,b\right)=\left(0.3,0.8\right),\left(0,0.5\right),\left(0.5,1.\right)$ for, respectively, the terms 'most', 'almost all', 'at least half' \cite{yager1988ordered,zhou2008type}. 

\end{definition}

\section{Driving Style Recognition}
\label{sec:driving_style_recognition}

This paper proposes a driving style detection using Mamdani type-2 Fuzzy Inference System (T2-FIS) with Interval Type-2 Fuzzy Sets and Fuzzy Multiple Experts Decision-Making (FMEDM) using Ordered Weighted Averaging (OWA) aggregate operator. A FIS is a rule-based decision-making technique that is able to handle uncertainties and subjectivities using fuzzy sets and fuzzy logic, and to represent experts' knowledge using a rulebase. Moreover, it is important to build the rules of the system using a multiple expert approach to increase the robustness of the system in the face of expert bias and other inaccuracies associated with the creation of the knowledge base. 

A driving style can be characterized by several features \cite{sagberg2015review,han2018statistical,martinez2017driving}. According to Martinez et al. \cite{martinez2017driving}, the combination of longitudinal and lateral features allows the detection of many vehicle movement patterns that characterize a particular driving style. The developed T2-FIS receives as input statistical measures (i.e., mean and standard deviation) of speed, acceleration, deceleration, and lateral jerk estimated over a time window. The output is one of driving style classes, i.e., \emph{calm}, \emph{moderate}, and \emph{aggressive}. The following sections explain the input features and their fuzzy representations (see Section \ref{subsec:method:features}); the T2-FIS model and the fuzzy representation of the driving styles (see Section \ref{subsec:method:fis}); and, the multiple-experts aggregation approach using the OWA operator (see Section \ref{subsec:method:fmedm}).

\subsection{Features}
\label{subsec:method:features}


The proposed T2-FIS has four input features that represent longitudinal and lateral kinematics parameters of the vehicle movement. 
Type-2 Interval Fuzzy sets with a trapezoidal membership function (see Definition \ref{def:fuzzy:trapmf}) model the input features (antecessors) and the output (consequent) of the system.  
Table \ref{tab:type2_fis:parameters} shows the parameters of the antecessors and consequent type-2 fuzzy sets. All the values of each fuzzy subset were defined by the joint analysis of the hierarchical discretization of the data, by the descriptive statistical analysis of the data (i.e., quantile, minimum value, maximum value, mean, and standard deviation), and by the analysis of the sets and centroids found in the existing works in the literature \cite{aljaafreh2012driving,dorr2014online,constantinescu2010driving,liu2019research,meiring2015review}. Moreover, trapezoidal membership functions represent all fuzzy numbers because they can model linguistic terms that represent interval values, and they are also simpler to analyze compared to other membership functions \cite{wu2019recommendations,wu2012twelve}.

\begin{table}[!tb]
\caption{Type-2 Fuzzy Sets Parameters}
\label{tab:type2_fis:parameters}
\centering
\resizebox{\textwidth}{!}{
\scriptsize

\begin{tabular}{cccrrrrr}
\hline
\textbf{Fuzzy Set}                                                                                                        & \textbf{Fuzzy Subset}                & \textbf{Bound} & \multicolumn{1}{c}{$\mathbf{a_1}$} & \multicolumn{1}{c}{$\mathbf{a_2}$} & \multicolumn{1}{c}{$\mathbf{a_3}$} & \multicolumn{1}{c}{$\mathbf{a_4}$} & \multicolumn{1}{c}{$\mathbf{h}$} \\ \hline
\multirow{10}{*}{\textbf{\begin{tabular}[c]{@{}c@{}}\\\\\\Mean  Velocity ($Km/h$)\end{tabular}}}                                       & \multirow{2}{*}{\textit{Very Slow}}  & \textit{UMF}   & 0                               & 0                               & 15                              & 25                              & 1                              \\
                                                                                                                          &                                      & \textit{LMF}   & 0                               & 0                               & 12                              & 20                              & 0.8                            \\ \cline{2-8} 
                                                                                                                          & \multirow{2}{*}{\textit{Slow}}       & \textit{UMF}   & 15                              & 25                              & 35                              & 45                              & 1                              \\
                                                                                                                          &                                      & \textit{LMF}   & 20                              & 28                              & 32                              & 40                              & 0.8                            \\ \cline{2-8} 
                                                                                                                          & \multirow{2}{*}{\textit{Normal}}     & \textit{UMF}   & 35                              & 45                              & 55                              & 65                              & 1                              \\
                                                                                                                          &                                      & \textit{LMF}   & 40                              & 48                              & 52                              & 60                              & 0.8                            \\ \cline{2-8} 
                                                                                                                          & \multirow{2}{*}{\textit{Fast}}       & \textit{UMF}   & 55                              & 65                              & 75                              & 85                              & 1                              \\
                                                                                                                          &                                      & \textit{LMF}   & 60                              & 68                              & 72                              & 80                              & 0.8                            \\ \cline{2-8} 
                                                                                                                          & \multirow{2}{*}{\textit{Very Fast}}  & \textit{UMF}   & 75                              & 85                              & 100                             & 100                             & 1                              \\
                                                                                                                          &                                      & \textit{LMF}   & 80                              & 88                              & 100                             & 100                             & 0.8                            \\ \hline
\multirow{6}{*}{\textbf{\begin{tabular}[c]{@{}c@{}}\\\\Mean Acceleration,\\  Mean  Deceleration\\($m/s^2$)\end{tabular}}} & \multirow{2}{*}{\textit{Small}}      & \textit{UMF}   & 0                               & 0                               & 2                               & 3.5                             & 1                              \\
                                                                                                                          &                                      & \textit{LMF}   & 0                               & 0                               & 1.5                             & 2.7                             & 0.8                            \\ \cline{2-8} 
                                                                                                                          & \multirow{2}{*}{\textit{Medium}}     & \textit{UMF}   & 2                               & 3.5                             & 5.5                             & 7                               & 1                              \\
                                                                                                                          &                                      & \textit{LMF}   & 2.7                             & 4                               & 5                               & 6.2                             & 0.8                            \\ \cline{2-8} 
                                                                                                                          & \multirow{2}{*}{\textit{Large}}      & \textit{UMF}   & 5.5                             & 7                               & 10                              & 10                              & 1                              \\
                                                                                                                          &                                      & \textit{LMF}   & 6.2                             & 7.5                             & 10                              & 10                              & 0.8                            \\ \hline
\multirow{6}{*}{\textbf{\begin{tabular}[c]{@{}c@{}}\\STD Lateral  Jerk ($m/s^3$)\end{tabular}}}                                   & \multirow{2}{*}{\textit{Small}}      & \textit{UMF}   & 0                               & 0                               & 4                               & 6                               & 1                              \\
                                                                                                                          &                                      & \textit{LMF}   & 0                               & 0                               & 3                               & 4.5                             & 0.8                            \\ \cline{2-8} 
                                                                                                                          & \multirow{2}{*}{\textit{Medium}}     & \textit{UMF}   & 3                               & 6                               & 10                              & 12.9                            & 1                              \\
                                                                                                                          &                                      & \textit{LMF}   & 4.5                             & 6.9                             & 9.1                             & 11.4                            & 0.8                            \\ \cline{2-8} 
                                                                                                                          & \multirow{2}{*}{\textit{Large}}      & \textit{UMF}   & 10                              & 12                              & 16                              & 16                              & 1                              \\
                                                                                                                          &                                      & \textit{LMF}   & 11.5                            & 13                              & 16                              & 16                              & 0.8                            \\ \hline
\multirow{6}{*}{\textbf{\begin{tabular}[c]{@{}c@{}}\\Driving Style\end{tabular}}}                                         & \multirow{2}{*}{\textit{Calm}}       & \textit{UMF}   & 0                               & 0                               & 0.2                             & 0.4                             & 1                              \\
                                                                                                                          &                                      & \textit{LMF}   & 0                               & 0                               & 0.15                            & 0.3                             & 0.8                            \\ \cline{2-8} 
                                                                                                                          & \multirow{2}{*}{\textit{Moderate}}   & \textit{UMF}   & 0.2                             & 0.4                             & 0.6                             & 0.8                             & 1                              \\
                                                                                                                          &                                      & \textit{LMF}   & 0.3                             & 0.45                            & 0.55                            & 0.7                             & 0.8                            \\ \cline{2-8} 
                                                                                                                          & \multirow{2}{*}{\textit{Aggressive}} & \textit{UMF}   & 0.6                             & 0.8                             & 1                               & 1                               & 1                              \\
                                                                                                                          &                                      & \textit{LMF}   & 0.7                             & 0.85                            & 1                               & 1                               & 0.8                            \\ \hline
\end{tabular}
}

{\vspace{1em} \scriptsize UMF - Upper Membership Function; LMF - Lower Membership Function.}
\end{table}

\begin{itemize}
    \item \textbf{Mean Velocity} ($\bar v$): the first entry of the T2-FIS is the mean of the vehicle velocity over a time window. This measure is discretized by 5 type-2 fuzzy subsets using the trapezoidal membership function. As stated in  Han et al. \cite{han2018statistical}, and  Liu et al. \cite{liu2019research}, aggressive drivers usually has a high velocity profile, therefore, this longitudinal features are relevant in distinguishing driving styles.
 
    \item \textbf{Mean Acceleration} ($\bar a_{+}$) and \textbf{Mean Deceleration} ($\bar a_{-}$): the second and third input features are the mean of the positive acceleration ($\bar a_{+}$) and negative acceleration ($\bar a_{-}$), which occurs when the vehicle increases or decreases its velocity, respectively. These features are also relevant for driving style classification, since they measure how the vehicle speeds up and slows down. According to Wang and Lukic \cite{wang2011review} and  Lv et al. \cite{lv2018driving}, the average acceleration and deceleration of aggressive drivers are greater than moderate drivers, which, in turn, are greater than calm drivers. These entries have the same discretization, with 3 fuzzy subsets within the range $[0,10]$.
    
    \item \textbf{STD Lateral Jerk} ($\sigma_{j_y}$): jerk is the derivative of acceleration, that is, it is a  physical measure that quantifies the change in acceleration. As reported by Murphey, Milton and Kiliaris \cite{murphey2009driver}, the jerk profile  demonstrates how the driver changes his acceleration, so it can be used to define aggressiveness, in which, aggressive drivers have higher jerk profile. Similarly, Zylius \cite{zylius2017investigation} showed that the standard deviation is also higher for aggressive driving. Therefore, in this paper, the fourth entry is the standard deviation of the lateral jerk profile, which characterized how the driver changes its lateral acceleration. This entry is discretized by 3 fuzzy subsets within the range $[0,16]$. 
\end{itemize}

\subsection{Type-2 Fuzzy Inference System}
\label{subsec:method:fis}

In general, a T2-FIS consists of five components, as described in Section \ref{subsec:fuzzy2:fls}. The proposed T2-FIS combines lateral and longitudinal features into four type- 2 antecessors sets (see Section \ref{subsec:method:features}), namely: \emph{Mean Velocity}, with five subsets; \emph{Mean Acceleration}, with three subsets; \emph{Mean Deceleration}, with three subsets; and, \emph{STD Lateral Jerk}, with three subsets. 

The output of the T2-FIS, also called consequent, is the fuzzy set `\emph{Driving Style}' with three subsets with trapezoidal membership functions: \emph{calm}, characterized by a low kinematic profile, as drivers adopt a more conservative and cautious driving style; \emph{moderate}, with medium kinematic profiles; and, \emph{aggressive}, characterized by high kinematic profiles, such as high speed, acceleration, and variation of lateral jerk, as drivers take more risks while driving and tend to commit more traffic violations or perform riskier maneuvers. Table \ref{tab:type2_fis:parameters}  shows the parameters of each subset. 

The rulebase was created using the multiple expert decision-making, described in Section \ref{subsec:method:fmedm}, with the judgment of $n$ experts ($n=8$). Each expert was responsible for judging the 135 rules resulting from the combination of all entries. The output of each fuzzy rule is also a fuzzy set (one of the fuzzy subsets of the \emph{driving style}). The inference engine uses the  \emph{minimum t-norm} operator for reasoning and \emph{center-of-sets} as a method for type-reduction. Finally, defuzzification is performed by computing the centroid of the fuzzy set output.

\subsection{Multiple Expert Decision-Making}
\label{subsec:method:fmedm}

A Fuzzy Multiple Expert Decision-Making (FMEDM) combines the opinion of $n$ experts to build the rulebase of the system, which is also the structure that represents the knowledge used for decision-making. According to Vaníček, Vrana and Aly \cite{vanivcek2009fuzzy}, before the aggregation, the opinions' experts must be fuzzified in the same universe of discourse with domain within the interval $[0,1]$. In this paper, we use nine linguistic terms, showed in Table \ref{tab:multiple_expert_decision}, 
that may be chosen by the experts to judge each rule. Each linguistic term is mapped into the domain using the function $\mathit{f}(x)=\frac{x}{9}$, and the aggregation result (a value within the range $[0,1]$) is fuzzified into the `\emph{Driving Style}' fuzzy set.

\begin{table}[!tb]
    \caption{Linguist Terms of the Fuzzy Multiple Expert Decision-Making}
    \label{tab:multiple_expert_decision}
    \begin{center}
        \begin{tabular}{cc}
            \hline
            \textbf{Linguistic Term} & \textbf{Number} ($x$)\\
            \hline
            \textit{Calm} & 1\\
            \hline
            \textit{More Calm Than Moderate} & 2\\
            \hline
            \textit{Between Calm And Moderate} & 3\\
            \hline
            \textit{More Moderate Than Calm} & 4\\
            \hline
            \textit{Moderate} & 5\\
            \hline
            \textit{More Moderate Than Aggressive} & 6\\
            \hline
            \textit{Between Moderate And Aggressive} & 7\\
            \hline
            \textit{More Aggressive Than Moderate} & 8\\
            \hline
            \textit{Aggressive} & 9\\
            \hline
        \end{tabular}
    \end{center}
\end{table}

The combination process uses a fuzzy aggregation operator,  with either a fuzzy set or crisp values as output. In this paper, we use the Ordered Weighted Averaging (OWA) with linguistic quantifier, whose parameters $(a,b)=(0, 0.5)$ \cite{yager1988ordered,zhou2008type}. 
This linguistic quantifier gives preference to the highest values, or decisions, made by the experts for a given rule. In the proposed approach, the highest values comprise the most aggressive driving style, thus, in the process of aggregating the opinions of experts, opinions closer to the \emph{aggressive} style are given higher weight than those closest to the \emph{calm} style, giving a more conservative connotation to the decision system.

\section{Experimental Results and Discussion}
\label{sec:results}

To evaluate the proposed T2-FIS, we performed a descriptive statistical analysis of the groups found as a result of the classification, and compared them with four clustering algorithms, in addition to a T1-FIS created using only the upper membership function of the type-2 fuzzy sets (see Table \ref{tab:type2_fis:parameters}). The data are from the public dataset called Argoverse $v.1$ \cite{chang2019argoverse}.
The implementations of the clustering algorithms, the type-1 and type-2 fuzzy systems, and the extraction and preprocessing of the data are available in a public repository on GitHub\footnote{\url{https://github.com/iag0g0mes/t2fis_driving_style.git}}. The following sections describe in detail the dataset and preprocessing steps, the evaluation metrics and benchmark techniques, and the experimental results.

\subsection{Dataset}
\label{subsec:results:datasets}

The Argoverse dataset was developed by the North American company Argo AI, in collaboration with Carnegie Mellon and Georgia Institute of Technology universities. The database contains data from various sensors and semantic maps of the environment in several American cities. It also provides support for 3D object tracking and path prediction. It contains information from more than ten thousand tracked objects, including trajectories at intersections, lane changes, congestion, and curves \cite{chang2019argoverse}.

In this paper, we use the motion forecasting dataset with $324\;557$ scenarios, each $5$ seconds long, and divided into sample, training, testing, and validation. We extracted the features (i.e., mean velocity, mean acceleration, mean deceleration, standard deviation of lateral jerk) from the training and validation datasets, with $205\;942$ and $39\;472$ samples, respectively. The size of the window used to measure the features was 5 seconds. The training data were used to build all models, and the validation data were used for evaluation. Table \ref{tab:dataset:val} shows the descriptive statistical analysis of the validation data.

 \begin{table}[!bt]
 
       
    \caption{Descriptive Statistical Analysis of the Argoverse Dataset - Validation}
    \label{tab:dataset:val}
     \resizebox{\textwidth}{!}{
      \centering
        \begin{tabular}{clrrrrrrr}
\hline
\textbf{Filter}                         & \multicolumn{1}{c}{\textbf{Feature}} & \multicolumn{1}{c}{\textbf{Mean}} & \multicolumn{1}{c}{\textbf{STD}} & \multicolumn{1}{c}{\textbf{Min}} & \multicolumn{1}{c}{\textbf{Max}} & \multicolumn{1}{c}{\textbf{Q25}} & \multicolumn{1}{c}{\textbf{Q50}} & \multicolumn{1}{c}{\textbf{Q75}} \\ \hline
\multirow{4}{*}{\textbf{Raw}}           & \textit{Mean Velocity}               & \textit{9.46}                     & \textit{4.26}                    & 1.77                             & 24.27                            & 5.96                             & 8.9                              & 12.5                             \\
                                        & Mean Acceleration                    & 36.63                             & 22.67                            & 2.23                             & 310.85                           & 20.76                            & 30.47                            & 46.57                            \\
                                        & Mean Deceleration                    & 36.48                             & 22.47                            & 2.2                              & 214.87                           & 20.56                            & 30.39                            & 46.53                            \\
                                        & STD Lateral Jerk                     & 703.05                            & 973.48                           & 9.33                             & 51263.81                         & 307.67                           & 516.48                           & 897.95                           \\ \hline
\multirow{4}{*}{\textbf{EKF}}           & \textit{Mean Velocity}               & 9.33                              & 4.32                             & 1.29                             & 24.76                            & 5.8                              & 8.77                             & 12.43                            \\
                                        & \textit{Mean Acceleration}           & \textit{6.32}                     & 3.91                             & 0.49                             & 38.76                            & 3.55                             & 5.27                             & 8.08                             \\
                                        & \textit{Mean Deceleration}           & \textit{6.01}                     & 3.63                             & 0.37                             & 50.63                            & 3.39                             & 5.03                             & 7.73                             \\
                                        & \textit{STD Lateral Jerk}            & \textit{73.13}                    & 56                               & 1.44                             & 983.53                           & 33.56                            & 56.82                            & 97.8                             \\ \hline
\multirow{4}{*}{\textbf{SavitzkyGolay}} & \textit{Mean Velocity}               & \textit{9.68}                     & 4.35                             & 1.12                             & 33.77                            & 6.13                             & 9.32                             & 12.81                            \\
                                        & \textit{Mean Acceleration}           & \textit{1.22}                     & 1.05                             & 0                                & 19.07                            & 0.53                             & 0.95                             & 1.61                             \\
                                        & \textit{Mean Deceleration}           & \textit{1.25}                     & 1.13                             & 0                                & 21.95                            & 0.51                             & 0.95                             & 1.64                             \\
                                        & \textit{STD Lateral Jerk}            & \textit{1.42}                     & 1.35                             & 0.03                             & 26.91                            & 0.6                              & 1.04                             & 1.78                             \\ \hline
\end{tabular}
    }
    
    {\vspace{1em} \scriptsize Q - Quantile}
\end{table}

 We estimate all input features using the space derivative, since the dataset only provides the position of the vehicles and the timestamp. Based on Table \ref{tab:dataset:val}, it is possible to highlight the large amount of noise when estimating the features from the \emph{raw} data, which leads to unrealistic results. Therefore, we applied two filters to reduce the noise: an Extended Kalman Filter (EKF) and a Savitzky-Golay filter (polynomial degree 3 and a window size of 10 samples). The noise reduction by the SavitzkyGolay filter was greater than that of the EKF because this technique uses polynomial interpolation to smooth the data. The EKF uses dynamic equations of vehicle motion in a cycle with prediction (using the equations) and correction (using the observations).

\subsection{Evaluation Metrics}
\label{subsec:results:evaluation_metrics}

There are some challenges in evaluating driving style recognition without labeled data, as most evaluation metrics measure how well a method performs compared to the ground truth of the data. Some related work uses either supervised metrics (e.g., confusion matrix, accuracy, etc.) or clustering analyzes with manually labeled data obtained either in simulation \cite{han2018statistical,dorr2014online,wang2017driving}  or from some drivers who participated in the research \cite{johnson2011driving,suzdaleva2018online,constantinescu2010driving, liu2019research, deng2017driving, bejani2018context, brombacher2017driving, ma2021driving}. However, the amount of data, the variety of traffic situations, and the representativeness of kinematic profiles are more limited than with public and large-scale datasets such as Argoverse. In addition, studies using private data make comparison with other studies in the literature difficult. Some articles use private datasets to qualitatively evaluate the methods and results \cite{aljaafreh2012driving,qi2015leveraging,constantinescu2010driving}, or use an external variable, such as fuel consumption, to interpret the results \cite{murphey2009driver,suzdaleva2018online}.

According to Greene, Cunningham, and Mayer \cite{greene2008unsupervised}, in the absence of labeled data, there is no clear definition of what constitutes proper clustering. Therefore, it is possible to assess the outcome using qualitative methods and quantitative metrics through internal validation of the clusters. This involves measuring how well the clusters are defined, e.g., how much the clusters overlap and how variable the individual groups are. Another way to assess the outcome is through descriptive statistical analysis, which summarizes the data distribution based on central tendency and measures of dispersion \cite{demaris2013summarizing}.

Therefore, the proposed T2-FIS with decision-making by multiple experts is evaluated by descriptive statistical analysis using mean as the measure of central tendency, standard deviation, minimum value, maximum value, and quantiles as the measure of dispersion. The analysis of descriptive statistics makes it possible to understand how the data distribution behaves in each driving style (e.g., \emph{calm}, \emph{moderate}, and \emph{aggressive}) being able, therefore, to show the tendencies of the kinematic parameters for each class. For example, the central tendencies of the parameters can be used to indicate which drivers drive more conservatively and more safely. On the other hand, the dispersion of the data in each class allows us to see the consistency of the data in terms of their central tendencies, as well as the influence of noise on each class.

Furthermore, the proposed system is also compared with the implementation of four clustering algorithms \cite{xu2015comprehensive} (K-Means, Agglomerative Hierarchical Clustering, Fuzzy C-Means Clustering, and Gaussian Mixture Clustering), and with a Type-1 Fuzzy Inference (T1-FIS) system created using the upper membership functions, as shown in Table \ref{tab:type2_fis:parameters}. The experiments with the T1-FIS and T2-FIS were conducted using both the single expert and multiple expert approaches, with the same single expert used for both T1-FIS and T2-FIS.

In addition to the descriptive statistics, we also evaluated the clustering algorithms using three clustering analysis metrics \cite{greene2008unsupervised}, namely: Silhouette Coefficient, which estimates a value within the range $[-1,1]$, indicating how well distributed the data are in each cluster, with values close to $1$ being the best indicator; Calinski-Harabasz Score, where a larger value indicates better defined clusters with a high degree of cluster separation; and, Davis-Bouldin Index, where a lower value indicates a clustering algorithm with more compact and clearly separated clusters.

\subsection{Driving Style Recognition}
\label{subsec:results:driving_style_recognition}

In order to evaluate the proposed T2-FIS with multiple-experts decision-making, we compare its results with clustering algorithms and a T1-FIS using descriptive statistics analysis. According to  DeMaris and Selman \cite{demaris2013summarizing}, descriptive statistics summarizes the data distribution, which allow us to understand how each of the driving styles behaves regarding the tendency of the kinematics features. 


\subsubsection{Clustering Algorithms}

Tables \ref{tab:result:clustering_1:descriptive_stats} and  \ref{tab:result:clustering_2:descriptive_stats} shows the results of the four clustering algorithms, i.e., K-Means, Gaussian Mixture Models (GMM) Clustering, Fuzzy C-Means, and, Agglomerative Hierarchical Clustering. The outputs of clustering algorithms are clusters parameters, such as centers of each cluster, and, the label of each cluster is assigned lastly from the analysis of an expert. From these tables, it is possible to highlight the conformity of each driving style with the expected kinematics profiles, that is, the \emph{calm} driving style has a lower profile than the \emph{moderate} style, which, in turn, is lower than the \emph{aggressive} style. 


\begin{table*}[!bt]
\caption{K-Means and GMM Clustering - Descriptive Statistics Analysis}
\label{tab:result:clustering_1:descriptive_stats}

 \resizebox{\textwidth}{!}{
\scriptsize
\centering
\begin{tabular}{cclrrrrrrrr}
\hline
\multirow{2}{*}{\textbf{Driving Style}} & \multirow{2}{*}{\textbf{Filter}}       & \multicolumn{1}{c}{\multirow{2}{*}{\textbf{Feature}}} & \multicolumn{4}{c}{KMeans}  & \multicolumn{4}{c}{GMM Clustering} \\ \cline{4-11} 
&& \multicolumn{1}{c}{}    & \multicolumn{1}{c}{Mean} & \multicolumn{1}{c}{STD} & \multicolumn{1}{c}{Min} & \multicolumn{1}{c|}{Max}     & \multicolumn{1}{c}{Mean} & \multicolumn{1}{c}{STD} & \multicolumn{1}{c}{Min} & \multicolumn{1}{c}{Max} \\ \hline
\multirow{8}{*}{\textbf{Calm}}& \multirow{4}{*}{\textit{\textbf{EKF}}} & \textit{Mean Velocity ($m/s$)}      & 8.256& 4.162 & 1.285 & \multicolumn{1}{r|}{24.021}  & 8.370& 4.410 & 1.370 & 24.021\\
&& \textit{Mean Acceleration ($m/s^{2}$)} & 4.108& 1.948 & 0.486 & \multicolumn{1}{r|}{31.374}  & 3.313& 1.007 & 0.486 & 5.999 \\
&& \textit{Mean Deceleration ($m/s^{2}$)} & 3.910& 1.815 & 0.373 & \multicolumn{1}{r|}{50.628}  & 3.188& 0.987 & 0.373 & 5.923 \\
&& \textit{STD Lateral Jerk ($m/s^{3}$)}  & 36.811 & 15.238& 1.437 & \multicolumn{1}{r|}{66.028}  & 34.162 & 14.034& 1.437 & 75.518\\ \cline{2-11} 
& \multirow{4}{*}{\textit{\textbf{SavitzkyGolay}}} & \textit{Mean Velocity ($m/s$)}      & 5.217& 1.500 & 1.119 & \multicolumn{1}{r|}{8.092}   & 5.829& 2.117 & 1.144 & 12.852\\
&& \textit{Mean Acceleration ($m/s^{2}$)} & 1.395& 0.946 & 0.000 & \multicolumn{1}{r|}{9.761}   & 1.553& 0.736 & 0.000 & 3.951 \\
&& \textit{Mean Deceleration ($m/s^{2}$)} & 1.391& 1.011 & 0.000 & \multicolumn{1}{r|}{12.082}  & 1.510& 0.750 & 0.000 & 3.923 \\
&& \textit{STD Lateral Jerk ($m/s^{3}$)}  & 1.325& 1.211 & 0.037 & \multicolumn{1}{r|}{15.216}  & 0.986& 0.559 & 0.037 & 2.800 \\ \hline
\multirow{8}{*}{\textbf{Moderate}}      & \multirow{4}{*}{\textit{\textbf{EKF}}} & \textit{Mean Velocity ($m/s$)}      & 10.539 & 4.100 & 1.437 & \multicolumn{1}{r|}{24.760}  & 9.501& 4.081 & 1.285 & 23.050\\
&& \textit{Mean Acceleration ($m/s^{2}$)} & 7.987& 2.569 & 1.182 & \multicolumn{1}{r|}{34.482}  & 6.589& 1.830 & 1.588 & 12.069\\
&& \textit{Mean Deceleration ($m/s^{2}$)} & 7.710& 2.544 & 2.233 & \multicolumn{1}{r|}{34.482}  & 6.229& 1.748 & 1.381 & 10.986\\
&& \textit{STD Lateral Jerk ($m/s^{3}$)}  & 98.842 & 23.254& 65.056& \multicolumn{1}{r|}{150.668} & 75.457 & 30.741& 2.795 & 164.630       \\ \cline{2-11} 
& \multirow{4}{*}{\textit{\textbf{SavitzkyGolay}}} & \textit{Mean Velocity ($m/s$)}      & 10.256 & 1.565 & 7.472 & \multicolumn{1}{r|}{13.590}  & 11.145 & 3.992 & 1.119 & 23.983\\
&& \textit{Mean Acceleration ($m/s^{2}$)} & 1.082& 0.919 & 0.000 & \multicolumn{1}{r|}{11.373}  & 0.748& 0.441 & 0.000 & 2.352 \\
&& \textit{Mean Deceleration ($m/s^{2}$)} & 1.108& 0.990 & 0.000 & \multicolumn{1}{r|}{13.420}  & 0.737& 0.451 & 0.000 & 2.324 \\
&& \textit{STD Lateral Jerk ($m/s^{3}$)}  & 1.356& 1.129 & 0.033 & \multicolumn{1}{r|}{11.349}  & 1.075& 0.678 & 0.033 & 3.568 \\ \hline
\multirow{8}{*}{\textbf{Aggressive}}    & \multirow{4}{*}{\textit{\textbf{EKF}}} & \textit{Mean Velocity ($m/s$)}      & 11.537 & 4.098 & 1.880 & \multicolumn{1}{r|}{22.959}  & 11.077 & 4.077 & 1.880 & 24.760\\
&& \textit{Mean Acceleration ($m/s^{2}$)} & 13.908 & 4.493 & 2.669 & \multicolumn{1}{r|}{38.759}  & 12.410 & 4.171 & 1.182 & 38.759\\
&& \textit{Mean Deceleration ($m/s^{2}$)} & 12.743 & 3.965 & 2.307 & \multicolumn{1}{r|}{35.312}  & 11.784 & 3.566 & 2.307 & 50.628\\
&& \textit{STD Lateral Jerk ($m/s^{3}$)}  & 203.352& 56.289& 149.873       & \multicolumn{1}{r|}{983.533} & 154.555& 69.107& 4.441 & 983.533       \\ \cline{2-11} 
& \multirow{4}{*}{\textit{\textbf{SavitzkyGolay}}} & \textit{Mean Velocity ($m/s$)}      & 15.752 & 1.693 & 9.139 & \multicolumn{1}{r|}{33.766}  & 10.740 & 4.519 & 1.249 & 33.766\\
&& \textit{Mean Acceleration ($m/s^{2}$)} & 1.175& 1.336 & 0.000 & \multicolumn{1}{r|}{19.074}  & 2.493& 1.734 & 0.000 & 19.074\\
&& \textit{Mean Deceleration ($m/s^{2}$)} & 1.272& 1.461 & 0.000 & \multicolumn{1}{r|}{21.951}  & 2.826& 1.787 & 0.000 & 21.951\\
&& \textit{STD Lateral Jerk ($m/s^{3}$)}  & 1.693& 1.791 & 0.096 & \multicolumn{1}{r|}{26.908}  & 3.590& 2.098 & 0.065 & 26.908\\ \hline
\end{tabular}
}

\end{table*}

\begin{table*}[!bt]
\caption{Fuzzy C-Means and Agglomerative Hierarchical Clustering - Descriptive Statistics Analysis}
\label{tab:result:clustering_2:descriptive_stats}

 \resizebox{\textwidth}{!}{
\scriptsize
\centering
\begin{tabular}{cclrrrrrrrr}
\hline
\multirow{2}{*}{\textbf{Driving Style}} & \multirow{2}{*}{\textbf{Filter}}& \multicolumn{1}{c}{\multirow{2}{*}{\textbf{Feature}}} & \multicolumn{4}{c}{Fuzzy C-Means} & \multicolumn{4}{c}{Agglomerative Hierarchical Clustering}      \\ \cline{4-11} 
&& \multicolumn{1}{c}{}& \multicolumn{1}{c}{Mean} & \multicolumn{1}{c}{STD} & \multicolumn{1}{c}{Min} & \multicolumn{1}{c|}{Max}     & \multicolumn{1}{c}{Mean} & \multicolumn{1}{c}{STD} & \multicolumn{1}{c}{Min} & \multicolumn{1}{c}{Max} \\ \hline
\multirow{8}{*}{\textbf{Calm}}& \multirow{4}{*}{\textit{\textbf{EKF}}} & \textit{Mean Velocity ($m/s$)}         & 8.056   & 4.148  & 1.398  & \multicolumn{1}{r|}{24.021}  & 8.484   & 4.177  & 1.285  & 24.021 \\
&& \textit{Mean Acceleration ($m/s^{2}$)}    & 3.854   & 1.862  & 0.486  & \multicolumn{1}{r|}{25.735}  & 4.451   & 2.109  & 0.486  & 34.482 \\
&& \textit{Mean Deceleration ($m/s^{2}$)}    & 3.664   & 1.734  & 0.373  & \multicolumn{1}{r|}{50.628}  & 4.239   & 1.971  & 0.373  & 50.628 \\
&& \textit{STD Lateral Jerk ($m/s^{3}$)}     & 32.635  & 12.562 & 1.437  & \multicolumn{1}{r|}{55.130}  & 42.359  & 19.275 & 1.437  & 84.191 \\ \cline{2-11} 
& \multirow{4}{*}{\textit{\textbf{SavitzkyGolay}}} & \textit{Mean Velocity ($m/s$)}         & 4.911   & 1.356  & 1.119  & \multicolumn{1}{r|}{7.457}   & 6.487   & 2.269  & 1.119  & 14.432 \\
&& \textit{Mean Acceleration ($m/s^{2}$)}    & 1.389   & 0.935  & 0.000  & \multicolumn{1}{r|}{9.761}   & 1.425   & 0.997  & 0.000  & 12.782 \\
&& \textit{Mean Deceleration ($m/s^{2}$)}    & 1.377   & 0.984  & 0.000  & \multicolumn{1}{r|}{11.058}  & 1.422   & 1.043  & 0.000  & 11.422 \\
&& \textit{STD Lateral Jerk ($m/s^{3}$)}     & 1.301   & 1.183  & 0.037  & \multicolumn{1}{r|}{14.720}  & 1.459   & 1.241  & 0.033  & 13.772 \\ \hline
\multirow{8}{*}{\textbf{Moderate}}      & \multirow{4}{*}{\textit{\textbf{EKF}}} & \textit{Mean Velocity ($m/s$)}         & 10.143  & 4.085  & 1.285  & \multicolumn{1}{r|}{24.760}  & 10.890  & 4.098  & 1.437  & 24.760 \\
&& \textit{Mean Acceleration ($m/s^{2}$)}    & 7.069   & 2.339  & 1.182  & \multicolumn{1}{r|}{34.482}  & 8.717   & 2.502  & 1.182  & 24.540 \\
&& \textit{Mean Deceleration ($m/s^{2}$)}    & 6.817   & 2.289  & 2.168  & \multicolumn{1}{r|}{24.760}  & 8.453   & 2.520  & 2.250  & 24.772 \\
&& \textit{STD Lateral Jerk ($m/s^{3}$)}     & 84.410  & 20.390 & 53.868 & \multicolumn{1}{r|}{127.332} & 110.511 & 19.554 & 78.929 & 155.889\\ \cline{2-11} 
& \multirow{4}{*}{\textit{\textbf{SavitzkyGolay}}} & \textit{Mean Velocity ($m/s$)}         & 9.256   & 1.317  & 6.901  & \multicolumn{1}{r|}{11.567}  & 11.412  & 1.121  & 8.726  & 13.989 \\
&& \textit{Mean Acceleration ($m/s^{2}$)}    & 1.182   & 0.986  & 0.000  & \multicolumn{1}{r|}{11.373}  & 0.707   & 0.456  & 0.000  & 3.899  \\
&& \textit{Mean Deceleration ($m/s^{2}$)}    & 1.209   & 1.075  & 0.000  & \multicolumn{1}{r|}{15.031}  & 0.724   & 0.497  & 0.000  & 3.433  \\
&& \textit{STD Lateral Jerk ($m/s^{3}$)}     & 1.394   & 1.225  & 0.033  & \multicolumn{1}{r|}{15.216}  & 0.969   & 0.641  & 0.046  & 3.814  \\ \hline
\multirow{8}{*}{\textbf{Aggressive}}    & \multirow{4}{*}{\textit{\textbf{EKF}}} & \textit{Mean Velocity ($m/s$)}         & 11.496  & 4.135  & 1.880  & \multicolumn{1}{r|}{22.959}  & 11.529  & 4.062  & 1.880  & 22.959 \\
&& \textit{Mean Acceleration ($m/s^{2}$)}    & 12.663  & 4.266  & 2.253  & \multicolumn{1}{r|}{38.759}  & 14.044  & 4.507  & 2.669  & 38.759 \\
&& \textit{Mean Deceleration ($m/s^{2}$)}    & 11.777  & 3.797  & 2.307  & \multicolumn{1}{r|}{35.312}  & 12.818  & 3.992  & 2.307  & 35.312 \\
&& \textit{STD Lateral Jerk ($m/s^{3}$)}     & 179.674 & 55.019 & 126.581& \multicolumn{1}{r|}{983.533} & 205.515 & 56.440 & 149.215& 983.533\\ \cline{2-11} 
& \multirow{4}{*}{\textit{\textbf{SavitzkyGolay}}} & \textit{Mean Velocity ($m/s$)}         & 14.713  & 2.355  & 10.179 & \multicolumn{1}{r|}{33.766}  & 15.740  & 2.140  & 6.837  & 33.766 \\
&& \textit{Mean Acceleration ($m/s^{2}$)}    & 1.095   & 1.188  & 0.000  & \multicolumn{1}{r|}{19.074}  & 1.204   & 1.362  & 0.000  & 19.074 \\
&& \textit{Mean Deceleration ($m/s^{2}$)}    & 1.174   & 1.299  & 0.000  & \multicolumn{1}{r|}{21.951}  & 1.334   & 1.542  & 0.000  & 21.951 \\
&& \textit{STD Lateral Jerk ($m/s^{3}$)}     & 1.572   & 1.587  & 0.064  & \multicolumn{1}{r|}{26.908}  & 1.762   & 1.873  & 0.096  & 26.908 \\ \hline
\end{tabular}
}

\end{table*}

Table \ref{tab:clustering_analysis} shows the results of the internal clustering evaluation metrics of the four clustering algorithms. As stated in  Greene, Cunningham and Mayer \cite{greene2008unsupervised}, the closer the value of the \emph{Silhouette Coefficient} is to 1 the better allocated the data are to their respective clusters. Thus, from Table \ref{tab:clustering_analysis}, the algorithms that have the best silhouette coefficient are the \emph{Agglomerative Hierarchical Clustering} (EKF) and \emph{K-Means} (SavitzkyGolay). However,  \emph{K-Means} clustering outperforms the other clustering algorithms in the other two evaluation metrics, i.e., \emph{Calinski-Harabasz Score} and \emph{Davis-Bouldin Index}.

Another important observation from the results presented in Table \ref{tab:clustering_analysis}  is that the clustering algorithms that used the data filtered by the EKF present more defined shapes and are more separated in the feature space. The \emph{Calinski-Harabasz Score} values are higher for all features in the EKF filter compared to the SavitzkyGolay, just as the \emph{Silhouette Coefficient} values are closer to 1 with the EKF data, and the \emph{Davis-Bouldin Index} values are higher for the data using the SavitzkyGolay filter.  This behavior can be explained by the low spread of the data (see Table \ref{tab:dataset:val}), measured by the standard deviation,  which shows that the data are closer in the feature space, making the separation of clusters more difficult.


\begin{table}[!bt]
    \caption{\label{tab:clustering_analysis}Clustering Analysis Results}
    
     \resizebox{\textwidth}{!}{
        \scriptsize
        \centering
\begin{tabular}{clrrr}
\hline
\textbf{Filter}        & \multicolumn{1}{c}{\textbf{\begin{tabular}[c]{@{}c@{}}Clustering \\ Algorithm\end{tabular}}} & \multicolumn{1}{c}{\textbf{\begin{tabular}[c]{@{}c@{}}Silhouette \\ Coefficient\end{tabular}}} & \multicolumn{1}{c}{\textbf{\begin{tabular}[c]{@{}c@{}}Calinski Harabasz \\ Score\end{tabular}}} & \multicolumn{1}{c}{\textbf{\begin{tabular}[c]{@{}c@{}}Davis-Bouldin \\ Index\end{tabular}}} \\ \hline
\multirow{4}{*}{\textbf{EKF}}          
      & \textit{K-Means}                  & 0.574      & \textbf{79194.484}   & \textbf{0.579}  \\
      & \textit{GMM-Clustering}           & 0.277      & 26441.647   & 0.946   \\
      & \textit{Fuzzy C-Means}            & 0.536      & 70604.611   & 0.605   \\
      & \begin{tabular}[l]{@{}l@{}}Agglomerative  Hierarchical Clustering\end{tabular}  & \textbf{0.580}      & 76792.967   & 0.582   \\ \hline
\multirow{4}{*}{\textbf{SavitzkyGolay}} 
      & \textit{K-Means}                  & \textbf{0.404}      & \textbf{46107.827}   & \textbf{0.850}   \\
      & \textit{GMM-Clustering}           & 0.160      & 8767.567    & 1.903   \\
      & \textit{Fuzzy C-Means}            & 0.371      & 43542.775   & 0.908   \\
      & \begin{tabular}[l]{@{}l@{}}Agglomerative  Hierarchical Clustering\end{tabular}  & 0.326      & 36235.493   & 0.871   \\ \hline
\end{tabular}
    }
\end{table}

\subsubsection{Type-1 and Type-2 Fuzzy Inference Systems}

\begin{table*}[!bt]
    \caption{Type-1 Fuzzy Inference System- Descriptive Statistics Analysis}
    \label{tab:fuzzy_result_1}
     \resizebox{\textwidth}{!}{
        \scriptsize
        \centering
\begin{tabular}{cclrrrrrrrr}
\hline
\multirow{2}{*}{\textbf{Driving Style}} & \multirow{2}{*}{\textbf{Filter}}       & \multicolumn{1}{c}{\multirow{2}{*}{\textbf{Feature}}} & \multicolumn{4}{c}{T1-FIS (Single)}    & \multicolumn{4}{c}{T1-FIS (Multiple)}       \\ \cline{4-11} 
 & & \multicolumn{1}{c}{}    & \multicolumn{1}{c}{Mean} & \multicolumn{1}{c}{STD} & \multicolumn{1}{c}{Min} & \multicolumn{1}{c|}{Max}     & \multicolumn{1}{c}{Mean} & \multicolumn{1}{c}{STD} & \multicolumn{1}{c}{Min} & \multicolumn{1}{c}{Max} \\ \hline
\multirow{8}{*}{\textbf{Calm}}& \multirow{4}{*}{\textit{\textbf{EKF}}} & \textit{Mean Velocity ($m/s$) }      & 4.606& 1.702         & 1.861         & \multicolumn{1}{r|}{9.860}   & 3.513& 0.941         & 1.861         & 8.374         \\
 & & \textit{Mean Acceleration ($m/s^{2}$) } & 1.798& 0.685         & 0.486         & \multicolumn{1}{r|}{5.474}   & 1.737& 0.640         & 0.518         & 3.270         \\
 & & \textit{Mean Deceleration ($m/s^{2}$) } & 1.709& 0.597         & 0.373         & \multicolumn{1}{r|}{3.694}   & 1.642& 0.581         & 0.373         & 3.269         \\
 & & \textit{STD Lateral Jerk ($m/s^{3}$) }  & 7.497& 2.201         & 1.437         & \multicolumn{1}{r|}{10.549}  & 7.157& 2.377         & 1.437         & 10.524        \\ \cline{2-11} 
 & \multirow{4}{*}{\textit{\textbf{SavitzkyGolay}}} & \textit{Mean Velocity ($m/s$) }      & 8.164& 3.222         & 1.119         & \multicolumn{1}{r|}{14.285}  & 5.713& 1.799         & 1.119         & 8.729         \\
 & & \textit{Mean Acceleration ($m/s^{2}$) } & 1.136& 0.784         & 0.000         & \multicolumn{1}{r|}{6.257}   & 1.237& 0.749         & 0.000         & 6.257         \\
 & & \textit{Mean Deceleration ($m/s^{2}$) } & 1.133& 0.808         & 0.000         & \multicolumn{1}{r|}{6.193}   & 1.203& 0.743         & 0.000         & 5.988         \\
 & & \textit{STD Lateral Jerk ($m/s^{3}$) }  & 1.237& 0.924         & 0.033         & \multicolumn{1}{r|}{9.974}   & 1.165& 0.861         & 0.037         & 9.974         \\ \hline
\multirow{8}{*}{\textbf{Moderate}}      & \multirow{4}{*}{\textit{\textbf{EKF}}} & \textit{Mean Velocity ($m/s$) }      & 6.798& 2.695         & 1.479         & \multicolumn{1}{r|}{14.949}  & 6.752& 3.004         & 1.479         & 14.949        \\
 & & \textit{Mean Acceleration ($m/s^{2}$) } & 3.672& 1.314         & 0.627         & \multicolumn{1}{r|}{8.168}   & 3.075& 0.984         & 0.486         & 8.168         \\
 & & \textit{Mean Deceleration ($m/s^{2}$) } & 3.391& 1.210         & 0.644         & \multicolumn{1}{r|}{7.609}   & 2.829& 0.793         & 0.644         & 7.609         \\
 & & \textit{STD Lateral Jerk ($m/s^{3}$) }  & 37.698         & 20.161        & 3.643         & \multicolumn{1}{r|}{192.741} & 30.350         & 13.789        & 2.795         & 142.778       \\ \cline{2-11} 
 & \multirow{4}{*}{\textit{\textbf{SavitzkyGolay}}} & \textit{Mean Velocity ($m/s$) }      & 15.107         & 3.282         & 1.916         & \multicolumn{1}{r|}{33.766}  & 12.463         & 3.082         & 1.916         & 20.539        \\
 & & \textit{Mean Acceleration ($m/s^{2}$) } & 1.357& 1.346         & 0.000         & \multicolumn{1}{r|}{12.782}  & 1.140& 1.038         & 0.000         & 12.782        \\
 & & \textit{Mean Deceleration ($m/s^{2}$) } & 1.493& 1.526         & 0.000         & \multicolumn{1}{r|}{13.698}  & 1.219& 1.167         & 0.000         & 13.698        \\
 & & \textit{STD Lateral Jerk ($m/s^{3}$) }  & 1.855& 1.730         & 0.101         & \multicolumn{1}{r|}{11.615}  & 1.520& 1.322         & 0.033         & 15.216        \\ \hline
\multirow{8}{*}{\textbf{Aggressive}}    & \multirow{4}{*}{\textit{\textbf{EKF}}} & \textit{Mean Velocity ($m/s$) }      & 11.002         & 4.344         & 1.285         & \multicolumn{1}{r|}{24.760}  & 10.436         & 4.303         & 1.285         & 24.760        \\
 & & \textit{Mean Acceleration ($m/s^{2}$) } & 8.062& 4.018         & 1.182         & \multicolumn{1}{r|}{38.759}  & 7.683& 3.853         & 1.182         & 38.759        \\
 & & \textit{Mean Deceleration ($m/s^{2}$) } & 7.723& 3.620         & 1.500         & \multicolumn{1}{r|}{50.628}  & 7.339& 3.507         & 1.500         & 50.628        \\
 & & \textit{STD Lateral Jerk ($m/s^{3}$) }  & 96.489         & 58.982        & 5.999         & \multicolumn{1}{r|}{983.533} & 91.106         & 57.001        & 5.999         & 983.533       \\ \cline{2-11} 
 & \multirow{4}{*}{\textit{\textbf{SavitzkyGolay}}} & \textit{Mean Velocity ($m/s$) }      & 16.903         & 4.084         & 4.113         & \multicolumn{1}{r|}{31.314}  & 18.159         & 4.796         & 4.392         & 33.766        \\
 & & \textit{Mean Acceleration ($m/s^{2}$) } & 4.587& 3.176         & 0.315         & \multicolumn{1}{r|}{19.074}  & 4.330& 3.339         & 0.203         & 19.074        \\
 & & \textit{Mean Deceleration ($m/s^{2}$) } & 5.113& 3.020         & 0.000         & \multicolumn{1}{r|}{21.951}  & 4.428& 3.551         & 0.000         & 21.951        \\
 & & \textit{STD Lateral Jerk ($m/s^{3}$) }  & 6.595& 3.921         & 0.388         & \multicolumn{1}{r|}{26.908}  & 5.947& 4.417         & 0.117         & 26.908        \\ \hline
\end{tabular}
    }
\end{table*}

\begin{table*}[!bt]
    \caption{Type-2 Fuzzy Inference System - Descriptive Statistics Analysis}
    \label{tab:fuzzy_result_2}
     \resizebox{\textwidth}{!}{
        \scriptsize
        \centering
\begin{tabular}{cclrrrrrrrr}
\hline
\multirow{2}{*}{\textbf{Driving Style}} & \multirow{2}{*}{\textbf{Filter}}       & \multicolumn{1}{c}{\multirow{2}{*}{\textbf{Feature}}} & \multicolumn{4}{c}{T2-FIS (Single)}    & \multicolumn{4}{c}{T2-FIS (Multiple)}       \\ \cline{4-11} 
 & & \multicolumn{1}{c}{}    & \multicolumn{1}{c}{Mean} & \multicolumn{1}{c}{STD} & \multicolumn{1}{c}{Min} & \multicolumn{1}{c|}{Max}     & \multicolumn{1}{c}{Mean} & \multicolumn{1}{c}{STD} & \multicolumn{1}{c}{Min} & \multicolumn{1}{c}{Max} \\ \hline
\multirow{8}{*}{\textbf{Calm}}& \multirow{4}{*}{\textit{\textbf{EKF}}} & \textit{Mean Velocity ($m/s$)}      & 4.802& 2.133         & 1.861         & \multicolumn{1}{r|}{10.790}  & 3.556& 0.950         & 1.861         & 8.374         \\
 & & \textit{Mean Acceleration ($m/s^{2}$)} & 1.551& 0.483         & 0.486         & \multicolumn{1}{r|}{2.996}   & 1.565& 0.511         & 0.486         & 2.996         \\
 & & \textit{Mean Deceleration ($m/s^{2}$)} & 1.502& 0.443         & 0.373         & \multicolumn{1}{r|}{2.876}   & 1.499& 0.457         & 0.373         & 2.537         \\
 & & \textit{STD Lateral Jerk ($m/s^{3}$)}  & 7.777& 2.322         & 1.437         & \multicolumn{1}{r|}{10.999}  & 7.464& 2.450         & 1.437         & 10.999        \\ \cline{2-11} 
 & \multirow{4}{*}{\textit{\textbf{SavitzkyGolay}}} & \textit{Mean Velocity ($m/s$)}      & 8.830& 3.706         & 1.119         & \multicolumn{1}{r|}{16.354}  & 6.673& 2.413         & 1.119         & 10.799        \\
 & & \textit{Mean Acceleration ($m/s^{2}$)} & 1.060& 0.726         & 0.000         & \multicolumn{1}{r|}{5.877}   & 1.086& 0.665         & 0.000         & 5.877         \\
 & & \textit{Mean Deceleration ($m/s^{2}$)} & 1.056& 0.751         & 0.000         & \multicolumn{1}{r|}{6.003}   & 1.059& 0.659         & 0.000         & 5.988         \\
 & & \textit{STD Lateral Jerk ($m/s^{3}$)}  & 1.173& 0.835         & 0.033         & \multicolumn{1}{r|}{7.897}   & 1.116& 0.794         & 0.033         & 6.996         \\ \hline
\multirow{8}{*}{\textbf{Moderate}}      & \multirow{4}{*}{\textit{\textbf{EKF}}} & \textit{Mean Velocity ($m/s$)}      & 7.520& 2.795         & 1.544         & \multicolumn{1}{r|}{17.750}  & 7.008& 3.425         & 1.544         & 17.750        \\
 & & \textit{Mean Acceleration ($m/s^{2}$)} & 3.713& 1.311         & 0.709         & \multicolumn{1}{r|}{8.168}   & 2.697& 0.858         & 0.627         & 8.168         \\
 & & \textit{Mean Deceleration ($m/s^{2}$)} & 3.475& 1.269         & 0.702         & \multicolumn{1}{r|}{8.359}   & 2.473& 0.689         & 0.644         & 6.811         \\
 & & \textit{STD Lateral Jerk ($m/s^{3}$)}  & 38.910         & 20.962        & 2.795         & \multicolumn{1}{r|}{192.741} & 25.660         & 11.402        & 2.696         & 168.585       \\ \cline{2-11} 
 & \multirow{4}{*}{\textit{\textbf{SavitzkyGolay}}} & \textit{Mean Velocity ($m/s$)}      & 13.788         & 4.712         & 1.441         & \multicolumn{1}{r|}{25.899}  & 12.813         & 3.524         & 1.441         & 22.916        \\
 & & \textit{Mean Acceleration ($m/s^{2}$)} & 1.873& 1.528         & 0.000         & \multicolumn{1}{r|}{19.074}  & 1.287& 1.138         & 0.000         & 11.373        \\
 & & \textit{Mean Deceleration ($m/s^{2}$)} & 2.082& 1.677         & 0.000         & \multicolumn{1}{r|}{13.698}  & 1.385& 1.277         & 0.000         & 13.698        \\
 & & \textit{STD Lateral Jerk ($m/s^{3}$)}  & 2.465& 1.966         & 0.065         & \multicolumn{1}{r|}{13.966}  & 1.650& 1.429         & 0.046         & 15.216        \\ \hline
\multirow{8}{*}{\textbf{Aggressive}}    & \multirow{4}{*}{\textit{\textbf{EKF}}} & \textit{Mean Velocity ($m/s$)}      & 10.483         & 4.660         & 1.285         & \multicolumn{1}{r|}{24.760}  & 9.923& 4.307         & 1.285         & 24.760        \\
 & & \textit{Mean Acceleration ($m/s^{2}$)} & 7.949& 4.052         & 1.182         & \multicolumn{1}{r|}{38.759}  & 7.204& 3.837         & 1.182         & 38.759        \\
 & & \textit{Mean Deceleration ($m/s^{2}$)} & 7.586& 3.680         & 1.554         & \multicolumn{1}{r|}{50.628}  & 6.867& 3.520         & 1.554         & 50.628        \\
 & & \textit{STD Lateral Jerk ($m/s^{3}$)}  & 94.507         & 59.614        & 5.414         & \multicolumn{1}{r|}{983.533} & 84.703         & 56.245        & 5.414         & 983.533       \\ \cline{2-11} 
 & \multirow{4}{*}{\textit{\textbf{SavitzkyGolay}}} & \textit{Mean Velocity ($m/s$)}      & 17.718         & 4.645         & 4.113         & \multicolumn{1}{r|}{33.766}  & 17.037         & 5.393         & 5.118         & 33.766        \\
 & & \textit{Mean Acceleration ($m/s^{2}$)} & 4.781& 3.189         & 0.417         & \multicolumn{1}{r|}{17.386}  & 5.810& 3.117         & 0.492         & 19.074        \\
 & & \textit{Mean Deceleration ($m/s^{2}$)} & 5.084& 3.237         & 0.000         & \multicolumn{1}{r|}{21.951}  & 5.691& 3.450         & 0.000         & 21.951        \\
 & & \textit{STD Lateral Jerk ($m/s^{3}$)}  & 6.652& 4.173         & 0.507         & \multicolumn{1}{r|}{26.908}  & 7.748& 4.009         & 0.551         & 26.908        \\ \hline
\end{tabular}
    }
\end{table*}
This experiment has two parts, for both the T1-FIS (Table \ref{tab:fuzzy_result_1}) and the proposed T2-FIS (Table \ref{tab:fuzzy_result_2}). The first part uses the rulebase with a \emph{single} expert, and the second part uses \emph{multiple} experts to build the rulebase with aggregation of eight experts' opinions, $n=8$, (see Sections \ref{subsec:fuzzy2:multiple_experts_dm} and \ref{subsec:method:fmedm}).

Taking into consideration only the experiments with a \emph{single} expert,  the central tendencies of the \emph{calm} driving style are on average $\left\{4.70, 1.67, 1.61, 7.64\right\}_{EKF}$, and $\left\{8.5, 1.1, 1.09, 1.2\right\}_{SAVGOL}$. The \emph{moderate} driving style has means equal to $\left\{7.60, 3.69, 3.43, 38.30\right\}_{EKF}$ and $\left\{14.45, 1.62, 1.79, 2.16\right\}_{SAVGOL}$. In addition, the means of \emph{aggressive} driving style are  $\left\{10.74, 8.01, 7.65, 95.5\right\}_{EKF}$ and  $\left\{17.31, 4.68, 5.1, 6.62\right\}_{SAVGOL}$. The above values are expressed in the form of Equation \ref{eq:tuple}, whose notation we have also adopted in the remaining of this section.

\begin{eqnarray}
    \label{eq:tuple}
    \{\mu_{\bar v},\;\mu_{\bar a_{+}},\;\mu_{\bar a_{-}},\mu_{\sigma_{j_y}}\}_{FILTER}
\end{eqnarray}
where $\mu$ if the mean of the velocity ($\bar v$), acceleration ($a_{+}$), deceleration ($a_{-}$), and the STD of the lateral jerk ($\sigma_{j_{y}}$), across the mentioned experiment using the data from the $FILTER = \{EKF,\;SAVGOL\}$. In addition, $SAVGOL$ is the abbreviation of the SavitzkyGolay filter.

Alternatively, the results using the \emph{multiple} experts approaches showed that, the central tendencies of the \emph{calm} driving style are on average $\left\{3.53, 1.65, 1.57, 7.31\right\}_{EKF}$ and  $\left\{6.19, 1.16, 1.13, 1.14\right\}_{SAVGOL}$. In turn, the \emph{moderate} driving style has means equals to $\left\{6.88, 2.89, 2.65, 28.00\right\}_{EKF}$, and  $\left\{12.64, 1.21, 1.3, 1.58\right\}_{SAVGOL}$.  Moreover, the \emph{aggressive} driving style has means equals to $\left\{10.18, 7.44, 7.1, 87.9\right\}_{EKF}$, and $\left\{17.6, 5.07, 5.06, 6.85\right\}_{SAVGOL}$.

The central tendencies and standard deviations of the \emph{multiple} experts approaches are lower than the \emph{single} experts approaches for most of the features, in both the EKF and SavitzkyGolay filters.  These results show that the \emph{multiple} experts approaches find more distinct and regular kinematic profiles for each driving style, since spreads of the data from their central tendencies are lower.


In addition to the differences in the means and standard deviations of the data for each driving style between the expert approaches, the dispersion of the data as measured by the quantiles is also different. Table \ref{tab:q75_fuzzy} shows the average quantiles  $\mathcal{Q}_{0.75}$, also known as the third quartile ($\mathrm{Q}_{3}$), which is the upper limit of $75\%$ of the data, between the type-1 and type-2 fuzzy inference system. 
These quantile values show the lower kinematic tendency of the \emph{multiple} expert approach compared to the \emph{single} expert approach. The only exception is the \emph{aggressive} style when using the SavitzkyGolay filter, which occurred because some instances that were classified as \emph{aggressive} when using the \emph{multiple}  expert approach were classified as \emph{moderate} using only a \emph{single} expert. These differences can be seen in the maximum and minimum values in Tables \ref{tab:fuzzy_result_1} and  \ref{tab:fuzzy_result_2}. In addition, this exception also illustrates the difference in assessment between single or multiple experts, where in the latter case the kinematic profile of the driving style tends  to be lower.

\begin{table*}[!bt]
    \caption{Average $\mathcal{Q}_{0.75}$ between T1-FIS and T2-FIS}
    \label{tab:q75_fuzzy}
     \resizebox{\textwidth}{!}{
 \scriptsize
 \centering
\begin{tabular}{cccrrrr}
\hline
\multirow{2}{*}{\textbf{Driving Style}} & \multirow{2}{*}{\textbf{Filter}} & \multirow{2}{*}{\textbf{Approach}} & \multicolumn{4}{c}{\textbf{Features}}  \\ \cline{4-7} 
& &        & \multicolumn{1}{c}{\textit{\begin{tabular}[c]{@{}c@{}}Mean Velocity \\  ($m/s$)\end{tabular}}} & \multicolumn{1}{c}{\textit{\begin{tabular}[c]{@{}c@{}}Mean Acceleration \\  ($m/s^2$)\end{tabular}}} & \multicolumn{1}{c}{\textit{\begin{tabular}[c]{@{}c@{}}Mean Deceleration \\  ($m/s^2$)\end{tabular}}} & \multicolumn{1}{c}{\textit{\begin{tabular}[c]{@{}c@{}}STD Lateral Jerk \\  ($m/s^3$)\end{tabular}}} \\ \hline
\multirow{4}{*}{\textbf{Calm}}   & \multirow{2}{*}{\textit{\textbf{EKF}}}  & \textit{Single}      & 5.97 & 2.08      & 1.98      & 9.5      \\
& & \textit{Multiple}    & 4.21 & 2.09      & 1.94      & 9.35     \\ \cline{2-7} 
& \multirow{2}{*}{\textbf{SavitzkyGolay}} & \textit{Single}      & 11.24       & 1.53      & 1.52      & 1.6      \\
& & \textit{Multiple}    & 7.96 & 1.63      & 1.58      & 1.5      \\ \hline
\multirow{4}{*}{\textbf{Moderate}}      & \multirow{2}{*}{\textit{\textbf{EKF}}}  & \textit{Single}      & 8.88 & 4.65      & 4.23      & 49.19    \\
& & \textit{Multiple}    & 8.88 & \textit{3.33}    & 3.08      & 35.32    \\ \cline{2-7} 
& \multirow{2}{*}{\textbf{SavitzkyGolay}} & \textit{Single}      & 17.23       & 2.21      & 2.5       & 3.03     \\
& & \textit{Multiple}    & 14.98       & \textit{1.56}    & 1.74      & 2.05     \\ \hline
\multirow{4}{*}{\textbf{Aggressive}}    & \multirow{2}{*}{\textit{\textbf{EKF}}}  & \textit{Single}      & 14.23       & 9.92      & 9.49      & 123.38   \\
& & \textit{Multiple}    & 13.43       & 9.17      & 8.78      & 112.99   \\ \cline{2-7} 
& \multirow{2}{*}{\textbf{SavitzkyGolay}} & \textit{Single}      & 19.55       & 6.91      & 6.22      & 8.47     \\
& & \textit{Multiple}    & 21.14       & 7.19      & 7.11      & 8.98     \\ \hline
\end{tabular}
    }
\end{table*}

The descriptive statistics analysis also show differences between the T1-FIS and the T2-FIS using \emph{multiple} experts approach. 
The central tendencies of the features, `\emph{Mean Acceleration}' and `\emph{Mean Deceleration}', are lower in the T2-FIS than the T1-FIS, except for the \emph{moderate} and \emph{aggressive} driving styles using the data filtered with the SavitzkyGolay filter. Similarly, the maximum value for the two features in the results of the T2-FIS is less than or equal to the maximum values of the T1-FIS. Moreover, the interquartile ranges (IQR), Table \ref{tab:iqr_fuzzy}, of the `\emph{STD Lateral Jerk}', which has the highest amount of noise (as shown in Table \ref{tab:dataset:val}) show that its variation for each driving style class is lower with the T2-FIS. The only exception is the IQR of the \emph{moderate} driving style using SavitzkyGolay filter due to the difference of the quantiles $\mathcal{Q}_{0.75}$, which are $1.96$ (T1-FIS) and $2.14$ (T2-FIS).

\begin{table*}[!bt]
    \caption{Interquartile Ranges of T1-FIS and T2-FIS}
    \label{tab:iqr_fuzzy}
     \resizebox{\textwidth}{!}{
 \scriptsize
 \centering
 \begin{tabular}{cclrrrr}
\hline
\multirow{2}{*}{\textbf{Driving Style}} & \multirow{2}{*}{\textbf{Filter}}   & \multicolumn{1}{c}{\multirow{2}{*}{\textbf{Approach}}} & \multicolumn{4}{c}{\textbf{Features}} \\ \cline{4-7} 
  &         & \multicolumn{1}{c}{}       & \multicolumn{1}{c}{\textit{\begin{tabular}[c]{@{}c@{}}Mean Velocity \\ (m/s)\end{tabular}}} & \multicolumn{1}{c}{\textit{\begin{tabular}[c]{@{}c@{}}Mean Acceleration \\ (m/s2)\end{tabular}}} & \multicolumn{1}{c}{\textit{\begin{tabular}[c]{@{}c@{}}Mean Deceleration \\ (m/s2)\end{tabular}}} & \multicolumn{1}{c}{\textit{\begin{tabular}[c]{@{}c@{}}STD Lateral Jerk \\ (m/s3)\end{tabular}}} \\ \hline
\multirow{4}{*}{\textbf{Calm}}          & \multirow{2}{*}{\textit{\textbf{EKF}}}        & \textit{T1-FIS (Multiple)}    & \textbf{1.34}   & 0.94   & 0.8    & 3.78  \\
  &         & \textit{T2-FIS (Multiple)}    & 1.56 & \textbf{0.8}         & \textbf{0.69}        & \textbf{3.77}       \\ \cline{2-7} 
  & \multicolumn{1}{l}{\multirow{2}{*}{\textbf{SavitzkyGolay}}}          & \textit{T1-FIS (Multiple)}    & \textbf{2.88}   & 1.08   & 1.04   & 0.96  \\
  & \multicolumn{1}{l}{}  & \textit{T2-FIS (Multiple)}    & 3.99 & \textbf{0.97}         & \textbf{0.95}        & \textbf{0.93}       \\ \hline
\multirow{4}{*}{\textbf{Moderate}}      & \multirow{2}{*}{\textbf{EKF}}   & \textit{T1-FIS (Multiple)}    & \textbf{3.96}   & 1.19        & 1.06   & 17.97 \\
  &         & \textit{T2-FIS (Multiple)}    & 4.8  & \textbf{0.91}        & \textbf{0.77}        & \textbf{14.45}      \\ \cline{2-7} 
  & \multicolumn{1}{l}{\multirow{2}{*}{\textbf{SavitzkyGolay}}} & \textit{T1-FIS (Multiple)}    & 4.45 & \textbf{0.94}        & \textbf{1.09}        & \textbf{1.32}       \\
  & \multicolumn{1}{l}{}  & \textit{T2-FIS (Multiple)}    & \textbf{4.15}   & 1.22   & 1.46   & 1.47  \\ \hline
\multirow{4}{*}{\textbf{Aggressive}}    & \multirow{2}{*}{\textbf{EKF}}   & \textit{T1-FIS (Multiple)}    & 6.73 & \textbf{4.5}         & \textbf{4.25}        & 65.26 \\
  &         & \textit{T2-FIS (Multiple)}    & \textbf{6.7}    & 4.51   & 4.3    & \textbf{64.43}      \\ \cline{2-7} 
  & \multirow{2}{*}{\textbf{SavitzkyGolay}}       & \textit{T1-FIS (Multiple)}    & \textbf{6.18}   & 5.66   & 5.41   & 6.09  \\
  &         & \textit{T2-FIS (Multiple)}    & 7.57 & \textbf{4.01}        & \textbf{4.38}        & \textbf{4.4}        \\ \hline
\end{tabular}
    }
\end{table*}


\subsubsection{Type-2 Fuzzy Inference Systems and Clustering}

The differences of the descriptive statistics between the clustering algorithm and the T2-FIS using \emph{multiple} experts approach are also notable in the analysis of Tables \ref{tab:result:clustering_1:descriptive_stats} and  \ref{tab:fuzzy_result_2}. The contrast can also be noticed by the euclidean distance between the centers of the clusters from the K-Means (Table \ref{tab:centers_kmeans}),  and the central tendency of the T2-FIS. These distances are: \emph{calm}, $\{28.08, 9.65, 49.89\}_{EKF}$ and $\{1.72, 7.77, 14.91\}_{SAVGOL}$; \emph{moderate}, $\{89.41, 71.01, 11.61\}_{EKF}$ and $\{3.46, 2.72, 11.44\}_{SAVGOL}$; \emph{aggressive}, $\{197.62, 179.29, 119.91\}_{EKF}$ and $\{9.16, 3.08, 7.66\}_{SAVGOL}$. The above values are expressed in the form of Equation \ref{eq:tuple2}.

\begin{eqnarray}
    \label{eq:tuple2}
    \{d(x^{i}, y_{T2-FIS}^{calm}),\;d(x^{i}, y_{T2-FIS}^{moderate}),\;d(x^{i}, y_{T2-FIS}^{aggressive})\}_{FILTER}
\end{eqnarray}
where $d$ is the Euclidean distance, $x^{i}$ is the center or cluster of the K-Means with $i=\{calm,\;moderate,\;aggressive\}$, $y$ is the central tendency of the T2-FIS using multiple experts, and $FILTER=\{EKF, SAVGOL\}$.


\begin{table}[!bt]
    \caption{K-Means Center of Clusters}
    \label{tab:centers_kmeans}
     \resizebox{\textwidth}{!}{
        \scriptsize
        \centering
\begin{tabular}{ccrrrr}
\hline
\multirow{2}{*}{\textbf{Driving Style}} & \multirow{2}{*}{\textbf{Filter}} & \multicolumn{4}{c}{\textbf{Features}}     \\ \cline{3-6} 
  &     & \multicolumn{1}{c}{\textit{\begin{tabular}[c]{@{}c@{}}Mean Velocity \\ ($m/s$)\end{tabular}}} & \multicolumn{1}{c}{\textit{\begin{tabular}[c]{@{}c@{}}Mean Acceleration\\ ($m/s^2$)\end{tabular}}} & \multicolumn{1}{c}{\textit{\begin{tabular}[c]{@{}c@{}}Mean Deceleration\\ ($m/s^2$)\end{tabular}}} & \multicolumn{1}{c}{\textit{\begin{tabular}[c]{@{}c@{}}STD Lateral Jerk\\ ($m/s^3$)\end{tabular}}} \\ \hline
\multirow{2}{*}{\textbf{Calm}}          & \textbf{EKF}           & 7.21  & 4.27  & 3.99  & 35.06 \\
                                        & \textbf{SavitzkyGolay} & 5.05  & 1.48  & 1.42  & 1.30  \\ \hline
\multirow{2}{*}{\textbf{Moderate}}      & \textbf{EKF}           & 9.37  & 7.88  & 7.35  & 96.27 \\
                                        & \textbf{SavitzkyGolay} & 10.12 & 1.16  & 1.15  & 1.36  \\ \hline
\multirow{2}{*}{\textbf{Aggressive}}    & \textbf{EKF}           & 10.21 & 13.29 & 12.05 & 204.35\\
                                        & \textbf{SavitzkyGolay} & 15.64 & 1.86  & 1.85  & 2.62  \\ \hline
\end{tabular}
    }
\end{table}

These distances show that the center of the \emph{calm} cluster, using the EKF filter, is closer to the central tendency of the data classified as \emph{moderate} by the T2-FIS, whereas the center of the \emph{moderate} (K-Means) is closer to the central tendency of the \emph{aggressive} (T2-FIS). In turn, the center of the \emph{aggressive} style (K-Means), using the SavitzkyGolay filter, is closer to the \emph{moderate} style in the T2-FIS. Therefore, the lower kinematic profiles of the T2-FIS, using the data filtered with the EKF filter,   represent a more conservative classification. However, the central tendencies of the T2-FIS results using the SavitzkyGolay filter are higher than K-Means because the construction of the fuzzy intervals and rulebase follows a more general methodology, in contrast to the training of the K-Means algorithm that finds the centers of the clusters that best fit the current dataset.


The experimental results show that the fuzzy inference system found lower kinematic profiles and more consistent driving style classes with multiple expert decision-making. The T2-FIS also achieved lower data dispersion in each group, and lower central tendencies for the kinematics features \emph{Mean Acceleration} and \emph{Mean Deceleration} compared to the T1-FIS. In addition, the T2-FIS also behaves more conservatively than the clustering algorithms because its driving style classes have lower kinematic profiles for the data filtered with the EKF algorithm, which have the most noise.


\section{Conclusions}
\label{sec:conclusion}

This paper presents a driving style detection using a type-2 Fuzzy Inference System with Multiple Experts Decision-Making. 
Driving style summarizes a wide range of behaviors that are reflected in the kinematic parameters of the movement and may indicate a tendency for more violations, more risk in driving the vehicle, or higher fuel consumption, among others. Since there are many applications and meanings for driving style, the classification task is subject to a great deal of subjectivity and imprecision, in addition to the noise inherent in the input sensor readings. Therefore, type-2 fuzzy sets increase the ability of a fuzzy system to deal with imprecision, noise, and subjectivity, and the multiple expert approach reduces the effects of bias and other inaccuracies in the construction of the system's rulebase where all knowledge is stored.

The experimental results showed that the kinematic profiles of the driving styles classified by the fuzzy systems were lower, especially for the systems that used the multiple experts approach. One of the reasons that the multiple expert systems produced more conservative results was the heuristic used in aggregating the expert opinions, which assigned a higher weight to the opinions that were closest to the \emph{aggressive} driving style that was desirable for safety reasons.

Finally, the proposed methodology has room for improvement, such as: the use of type-2 fuzzy aggregation operators with the approach of multiple experts; the use of optimization algorithms, such as evolutionary algorithms, for the determination of the parameters of each fuzzy set \cite{wu2019recommendations}; and, the consideration of the temporal dependence between the transitions of the driving styles, such as the recurrent technique of Suzdaleva and Nagy \cite{suzdaleva2018online}.

\backmatter





\bmhead{Acknowledgments}

We thank the Coordination for the Improvement of Higher Education Personnel - Brazil (CAPES) for the financial support under grant 88887.500344/2020-0, the S\~ao Paulo Research Foundation (FAPESP) for the financial support under grant 2019/27301-7, and to all the experts who helped in the construction of the rulebase of the fuzzy system presented in this work.

\section*{Ethical Approval}

Not applicable

\section*{Consent to Participate}

Not applicable

\section*{Consent to Publish}

Not applicable

\section*{Authors Contributions}

All authors contributed to the study conception and design. Material preparation, data collection and analysis were performed by Iago Pach\^eco Gomes. The first draft of the manuscript was written by Iago Pach\^eco Gomes, and revised by Denis Fernando Wolf. All authors read and approved the final manuscript.

\section*{Funding}

This research was funded to the researcher Iago Pach\^eco Gomes, as a research grant, by the institutions:

\begin{itemize}
    \item \textit{Coordena\c c\~ao de Aperfei\c coamento de Pessoal de N\'ivel Superior - Brasil (CAPES)}: Finance Code 001 and grant 88887.500344/2020-0
    
    \item \textit{S\~ao Paulo Research Foundation (FAPESP)}: grant 2019/27301-7
\end{itemize}

\section*{Competing Interests}

Not applicable
\section*{Availability of data and materials}
Not applicable

\bibliography{references}


\end{document}